\documentclass[a4paper,fleqn]{cas-dc}

\usepackage[numbers,sort&compress]{natbib}
\usepackage{graphicx}
\usepackage{amssymb}
\usepackage{amsthm}
\usepackage{amsmath}
\usepackage{lineno}
\usepackage{enumitem}
\usepackage{multirow}
\usepackage{caption}
\usepackage{array}
\usepackage{booktabs}
\usepackage{tabularx}
\usepackage{longtable}
\usepackage{stfloats}
\usepackage{bm}
\usepackage{setspace}
\usepackage{natbib}
\usepackage{algorithm,algorithmicx,algpseudocode}
\usepackage{makecell}
\usepackage{color}
\usepackage{hyperref}
\usepackage{xcolor}
\def\tsc#1{\csdef{#1}{\textsc{\lowercase{#1}}\xspace}}
\tsc{WGM}
\tsc{QE}
\tsc{EP}
\tsc{PMS}
\tsc{BEC}
\tsc{DE}
\makeatletter
\newcommand{\multiline}[1]{%
  \begin{tabularx}{\dimexpr\linewidth-\ALG@thistlm}[t]{@{}X@{}}
    #1
  \end{tabularx}
}
\makeatother
\newcommand{\resultsfigwidth}{1.4in}
\newcommand{\resultscropwidth}{0.75in}
\newcommand{\resultsfigwidthab}{1.0in}
\newcommand{\resultscropwidthab}{0.9in}
\newcommand{\resultsfigwidthablego}{1.5in}
\newcommand{\resultscropwidthablego}{0.95in}

\newcommand{\cropflower}[2]{
  \makecell{
  \includegraphics[trim={390px 360px 498px 276px}, clip, width=\resultscropwidth]{#1} \\
  \includegraphics[trim={6px 636px 888px 6px}, clip, width=\resultscropwidth]{#1} \\
  {#2} \\
  }
}

\newcommand{\crophorn}[2]{
  \makecell{
  \includegraphics[trim={800px 270px 108px 386px}, clip, width=\resultscropwidth]{#1}   \\
  \includegraphics[trim={86px 556px 808px 86px}, clip, width=\resultscropwidth]{#1} \\
  {#2} \\
  }
}

\newcommand{\cropleaves}[2]{
  \makecell{
  \includegraphics[trim={600px 106px 278px 530px}, clip, width=\resultscropwidth]{#1} \\
  \includegraphics[trim={550px 506px 338px 130px}, clip, width=\resultscropwidth]{#1} \\
  {#2} \\
  }
}

\newcommand{\cropfern}[2]{
  \makecell{
  \includegraphics[trim={830px 270px 78px 386px}, clip, width=\resultscropwidth]{#1} \\
  \includegraphics[trim={550px 506px 338px 130px}, clip, width=\resultscropwidth]{#1} \\
  {#2} \\
  }
}

\newcommand{\crophotdog}[2]{
	\makecell{
		\includegraphics[trim={330px 450px 360px 250px}, clip, width=\resultscropwidth]{#1} \\
     \includegraphics[trim={400px 420px 290px 280px}, clip, width=\resultscropwidth]{#1}  \\
       {#2} \\
	}
}

\newcommand{\croplego}[2]{
  \makecell{
  \includegraphics[trim={340px 300px 356px 408px}, clip, width=\resultscropwidth]{#1} \\
   \includegraphics[trim={480px 450px 210px 250px}, clip, width=\resultscropwidth]{#1}  \\
       {#2} \\
  }
}
\newcommand{\croplegoab}[2]{
  \makecell{
  \includegraphics[trim={330px 450px 360px 250px}, clip, width=\resultscropwidthablego]{#1} \\
   \includegraphics[trim={170px 350px 551px 380px}, clip, width=\resultscropwidthablego]{#1}  \\
       {#2} \\
  }
}

\newcommand{\cropshipab}[2]{
  \makecell{
   \includegraphics[trim={400px 360px 260px 320px}, clip, width=\resultscropwidthab]{#1} \\
 \includegraphics[trim={120px 250px 500px 400px}, clip, width=\resultscropwidthab]{#1}  \\
       {#2} \\
  }
}

\newcommand{\cropship}[2]{
  \makecell{
  \includegraphics[trim={300px 410px 360px 270px}, clip, width=\resultscropwidth]{#1} \\
 \includegraphics[trim={120px 300px 500px 350px}, clip, width=\resultscropwidth]{#1}  \\
       {#2} \\
  }
}

\newcommand{\cropchair}[2]{
  \makecell{
  \includegraphics[trim={465px 297px 258px 426px}, clip, width=\resultscropwidth]{#1} \\
   \includegraphics[trim={280px 400px 420px 300px}, clip, width=\resultscropwidth]{#1}  \\
       {#2} \\
  }
}

\begin{document}
\let\printorcid\relax

\let\WriteBookmarks\relax
\def\floatpagepagefraction{1}
\def\textpagefraction{.001}

\shorttitle{ZS-SRT: An Efficient Zero-Shot Super-Resolution Training Method for Neural Radiance Fields}

\shortauthors{X. Feng, Y. He, Y. Wang, C. Wang et~al.}

\title [mode = title]{ZS-SRT: An Efficient Zero-Shot Super-Resolution Training Method for Neural Radiance Fields}                      



%
\author[1]{Xiang Feng}[type=editor,style=chinese]
\cormark[2]

\author[1]{Yongbo He}[type=editor,style=chinese]
\cormark[2]


\affiliation[1]{organization={School of Computer Science and Technology, Hangzhou Dianzi University},
    city={Hangzhou},
    postcode={310018}, 
    country={China}
}

\affiliation[fan]{organization={AI Lab at Lenovo Research},
            city={Beijing},
            country={China}}

\author[1]{Yubo Wang}[type=editor,style=chinese]

\author[1]{Chengkai Wang}[type=editor,style=chinese]

\author[1]{Zhenzhong Kuang}[type=editor,style=chinese]
\cormark[1]
\ead{zzkuang@hdu.edu.cn}

\author[1]{Jiajun Ding}[type=editor,style=chinese]
\author[1]{Feiwei Qin}[type=editor,style=chinese]
\author[fan]{Jianping Fan}[type=editor,style=chinese]

\cortext[cor1]{Corresponding Author: Zhenzhong Kuang.}
\cortext[cor2]{Equal Contribution.}



\begin{abstract}
Neural Radiance Fields (NeRF) have achieved great success in the task of synthesizing novel views that preserve the same resolution as the training views. However, it is challenging for NeRF to synthesize high-quality high-resolution novel views with low-resolution training data. To solve this problem, we propose a zero-shot super-resolution training framework for NeRF. This framework aims to guide the NeRF model to synthesize high-resolution novel views via single-scene internal learning rather than requiring any external high-resolution training data. Our approach consists of two stages. First, we learn a scene-specific degradation mapping by performing internal learning on a pretrained low-resolution coarse NeRF. Second, we optimize a super-resolution fine NeRF by conducting inverse rendering with our mapping function so as to backpropagate the gradients from low-resolution 2D space into the super-resolution 3D sampling space. Then, we further introduce a temporal ensemble strategy in the inference phase to compensate for the scene estimation errors. Our method is featured on two points: (1) it does not consume high-resolution views or additional scene data to train super-resolution NeRF; (2) it can speed up the training process by adopting a coarse-to-fine strategy. By conducting extensive experiments on public datasets, we have qualitatively and quantitatively demonstrated the effectiveness of our method.
\end{abstract}




\maketitle

\begin{figure*}[htb]
\centering
\includegraphics[width=\textwidth]{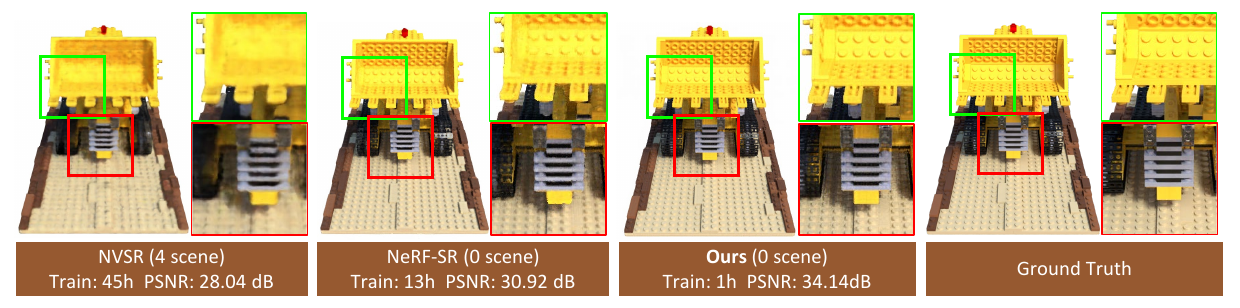}
\vspace{-0.6cm}
\caption{Comparing our method with existing super-resolution novel view synthesis methods. We have achieved optimal performance in three aspects: model training time, additional scene data consumed by model training, and the quality of synthesized novel views.}
\label{introduction}
\vspace{-0.4cm}
\end{figure*}

\section{Introduction}
Multi-view reconstruction is a long-standing problem in computer vision, with applications spanning virtual reality, augmented reality, telepresence, etc. Its goal is to estimate both scene appearance and geometry by harnessing information from multiple views. Over the past few decades, many traditional methods \cite{SFM,MVS} were proposed for this purpose. However, they often encounter multiple challenges, such as geometric structure estimation and texture recovery.

Recently, neural radiance fields (NeRF) \cite{NeRF} have attracted lots of attention for scene reconstruction based on neural inverse rendering and implicit representation. It's important to note that NeRF can render images at any desired resolution. However, NeRF struggles to generate super-resolution novel views effectively. This is because when it tries to render at a higher resolution than the training views, it encounters artifacts such as blurry due to the interpolation characteristics and the gap in sampling between training and testing. A straightforward way to solve this problem is to use high-resolution scene data to train a high-resolution NeRF. However, it is usually hard to obtain high-resolution scene data in many practical scenarios, such as the task of novel view synthesis and even text-to-3D content creation \cite{Magic3D,dreamfusion,make-it-3d}. Therefore, synthesizing high-resolution novel views without high-resolution ground truth is still challenging.

One possible solution to the above problem is to upsample the synthesized low-resolution novel views by using a well-trained single-image super-resolution (SISR) model \cite{ZSSR}. While the SISR model performs well on single images, it struggles to maintain multi-view consistency when processing multi-view images. In \cite{NVSR}, Bahat et al. proposed a neural volume super-resolution (NVSR) method to synthesize multi-view consistent high-resolution novel views, but it is time-consuming to train a generalizable model, and it is also difficult to apply NVSR due to the lack of scene data. In \cite{NeRF-SR}, Wang et al. proposed NeRF-SR by directly optimizing dense radiance fields through a super-sampling strategy to ensure view consistency. Although no additional scene data are required, NeRF-SR may lead to perceptual blurring and aliasing by enforcing the average value of high-resolution sub-pixels to match the value of low-resolution pixels.

In this paper, we propose a zero-shot super-resolution framework to improve the efficiency of reconstruction and generate reliable high-frequency details without consuming additional scene data. This two-stage framework integrates our idea of single-scene internal learning. First, we learn a scene-specific degradation mapping model on top of a pretrained coarse low-resolution NeRF. Second, we train a super-resolution NeRF in a coarse-to-fine manner to synthesize super-resolution novel views by conducting inverse rendering with the above well-trained degradation mapping model so as to backpropagate the gradients from low-resolution 2D space into the super-resolution 3D sampling space. Then, we further introduce a temporal ensemble strategy in the inference phase to compensate for the scene estimation errors. It is worth mentioning that our approach does not consume any high-resolution inputs and additional scene data during the training process and can ensure cross-view consistency. Besides, the employment of the coarse-to-fine strategy would help to speed up the training process. As shown in Figure \ref{introduction}, it is easy to observe that our approach outperforms NVSR \cite{NVSR} and NeRF-SR \cite{NeRF-SR} in terms of the training time, the amount of data required, and the synthetic quality.
In summary, our contributions are as follows:
\vspace{-0.2cm}
\begin{itemize}
\itemsep=0pt
    \item We propose a zero-shot and coarse-to-fine super-resolution training framework that only utilizes low-resolution ground truth to optimize the high-resolution 3D sampling space of radiance fields.
     \item We propose an internal learning method to obtain a scene-specific degradation mapping model which is further used in inverse rendering to optimize our fine NeRF in super-resolution training without consuming additional scene data.
    \item We propose a temporal ensemble strategy to compensate for the scene estimation errors.
    \item We qualitatively and quantitatively evaluate our method on a couple of public datasets, which has demonstrates the effectiveness of it.
\end{itemize}  

\section{Related work}

\subsection{Neural Radiance Fields}


Since implicit neural representation (INR) has obvious advantage in parameterizing various types of signals, it has been successfully applied in many recent works \cite{pifu,Neuralvolumes,siren,DeepSDF,SHI2024111145} to address problems that have plagued explicit representation for a long time. Unlike conventional discrete signal representation, INR relies on continuous function to describe various kinds of signals (e.g., image, audio, and 3D shape) by using neural networks to approximate the function for flexible and expressive representations. A straightforward advantage of employing continuous function lies in that it makes the signal representation no longer coupled with spatial resolution. As a result, the memory required for signal parameterization is independent of the spatial resolution and only scales with the complexity of the signal, indicating the infinite resolution of INR.


As a representative of INR, Neural Radiance Fields (NeRF) \cite{NeRF} receives increasing attention due to its excellent modeling ability on three-dimensional scene structure and surface characteristics (i.e. complex lighting and geometric effects) as well as the consistency across various viewpoints and lighting conditions. Initially, many variants of NeRF are proposed to address its inherent limitations. For example, some approaches \cite{DVGO,plenoctrees,NVSF,snerg} use voxel grids instead of MLPs to speed up training, and several other methods \cite{NoPe-NeRF,barf} reconstruct NeRF without camera poses. Subsequently, some works focus on generalizing NeRF to resolve challenges that have plagued other fields effectively. For example, NAF \cite{naf}, SNAF \cite{snaf}, and NeXF \cite{NeXF} explore new practical solutions to reconstruct medical objects by using sparse CBCT images. To further boost rendering quality, TensoRF \cite{TensoRF} proposes to factorize the 4D scene tensor into compact vector and matrix factors. In terms of multi-scale representation, Mip-NeRF \cite{Mip-NeRF} and Zip-NeRF \cite{Zip-NeRF} are proposed based on tapered sampling and adaptive position encoding to solve the aliasing problem caused by low-sampling rendering. Although NeRF and its variants can support the synthesis of novel views at any resolution, they suffer from difficulty in rendering high-quality, high-resolution views.

\subsection{Image Super-Resolution}

Single Image Super-Resolution (SISR) is a classic problem in computer vision that aims at recovering the lost details \cite{FENG2022109376,LIU2020106103}. In literature, there exist many supervised super-resolution methods, such as dense connections \cite{DenseNet,EDSR,rcan}, recursive structures \cite{DRCN,DRN}, back-projection \cite{dBPN}, and self-attention mechanisms \cite{Swinir,CAT,DAT}. Besides, the recent generative model is also exploited for image super-resolution by learning a mapping function from low resolution (LR) image to high resolution (HR) image \cite{SR3,ESRGAN,SRDiff,glean}.

The supervised super-resolution methods usually require paired training data, but this may not always be available in reality. Therefore, many researchers turn to explore self-supervised or unsupervised approaches. Most recently, Shocher et al. \cite{ZSSR} presents a zero-shot super-resolution (ZSSR) approach to produce high-quality super-resolution results by exploiting the internal recurrence of information inside a single image, where ZSSR does not require additional HR images. Although existing methods have achieved some success, they may confront the problem of maintaining multi-view consistency when dealing with multi-view images.

\subsection {Super-Resolution Novel View Synthesis}
To synthesize super-resolution novel views, many works focus on integrating the merits of both NeRF and SISR. In \cite{CGO-RF}, CGO-RF first relies on SISR to process multi-view images in advance and then trains a super-resolution update model with a large amount of high and low-resolution scene data to correct the spatial ambiguity caused by SISR. Similarly, NVSR \cite{NVSR} trains an EDSR model to super-resolve tri-planes. Since both of them need to learn generalization modules, they require large amounts of scene data pairs in the training process, which is also time-consuming. NeRF-SR \cite{NeRF-SR} attempts to recover high-frequency information in a sub-pixel way. However, it may confront the problem of blurry details by using the simple average pooling operation during the supervised training process of the model. In this paper, we address this problem by learning a degenerate mapping (from high-resolution radiance fields to low-resolution ground truth) on top of our proposed single-scene internal learning method so that we can generate super-resolution results with richer high-frequency details. 

\begin{figure*}
\label{pipeline}
\centering
\includegraphics[width=\textwidth]{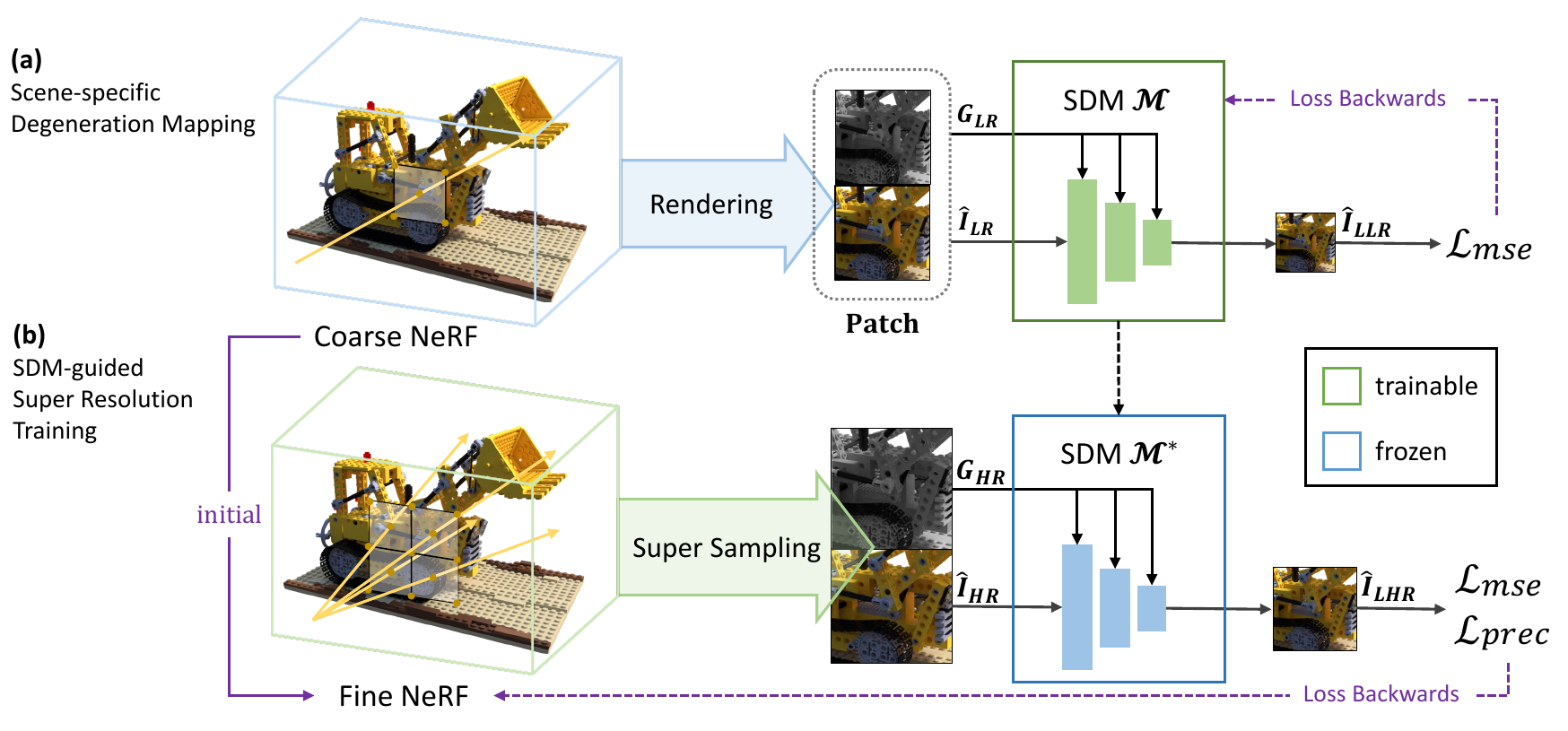}
\vspace{-0.6cm}
\caption{Overview of the proposed two-stage ZS-SRT framework for obtaining the high-resolution radiance field from low-resolution ground truth in a coarse-to-fine manner. In the first stage (a), a scene-specific degeneration mapping (SDM) model is learned on top of a coarsely NeRF. G denotes the gradient of the synthetic view, used to guide the training of the mapping relationship. In the second stage (b), a SDM-guided super-resolution training method is proposed to obtain a super-resolution fine NeRF so that our model can produce high-quality views with trusted, high-frequency details.}
\label{pipeline}
\vspace{-0.4cm}
\end{figure*}

\section{Preliminary: NeRF}

\textbf{Scene expression.} A radiance field is represented as a continuous function, $f$, which aims at mapping a 3D coordinate, $\mathbf{x} \in \mathbb{R}^3$, and a directional view unit vector, $\mathbf{d} \in \mathbb{S}^2$, to a volume density, $\sigma$, and $R G B$ values, $\mathbf{c}$. Typically, NeRF uses the following Multilayer Perception (MLP) to parameterize this function:
\begin{equation}
f_\theta:(\mathbf{x}, \mathbf{d}) \mapsto(\sigma, \mathbf{c})
\end{equation}
To improve the efficiency, Tensorial Radiance Fields (TensoRF) \cite{TensoRF} is adopted to realize NeRF, and its factorized definition is as follows:
\begin{equation}
\sigma, c=\sum_r \sum_m \mathcal{A}_{\sigma, r}^m(\mathbf{x}), S\left(\mathbf{B}\left(\oplus\left[\mathcal{A}_{c, r}^m(\mathbf{x})\right]_{m, r}\right), \mathbf{d}\right)
\end{equation}
Here, $\mathcal{A}_{\sigma, r}^m$ and $\mathcal{A}_{c, r}^m$ represent the factorized components for density and color, respectively.
The matrix $\mathbf{B}$ is used as a global appearance dictionary that abstracts
the appearance commonalities across the entire scene. Function $S$ processes the transformed color representations $\mathbf{B}(\oplus[\mathcal{A}_{c, r}^m(\mathbf{x})]_{m, r})$ alongside the viewing direction $\mathbf{d}$ to produce the final color $c$.

\textbf{Volume rendering.} For volume rendering, we employ differentiable volume rendering. For each pixel, we trace a ray $\mathbf{r}=\mathbf{o}+t \mathbf{d}$ by sampling a set of points and computing the pixel color as follows:
\vspace{-0.3cm}
\begin{equation}
\hat{C} = \sum_{i=1}^N \tau_i\left(1-\exp \left(-\sigma_i \Delta_i\right)\right) c_i
\end{equation}
\vspace{-0.5cm}
\begin{equation}
\tau_i = \exp \left(-\sum_{j=1}^{i-1} \sigma_j \Delta_j\right)
\end{equation}
where $\sigma_i$ and $c_i$ are the density and color computed at sampled locations $x_i ; \Delta_i$ is the ray step size, and $\tau_i$ denotes transmittance.

\textbf{Reconstruction.} To reconstruct a scene, given a set of multi-view input images with known camera poses, the tensorial radiance field per scene is optimized by minimizing the following photometric loss:
\begin{equation}
\mathcal{L}=\frac{1}{|\mathcal{R}|} \sum_{r \in \mathcal{R}}\|C(r)-\hat{C}(r)\|_2^2
\end{equation}
where $\mathcal{R}$ denotes the set of rays randomly sampled in each minibatch, $C(r)$ represents the ground truth pixel colors, and $\hat{C}(r)$ is the color computed through our model.

\section{Method}
Figure \ref{pipeline} illustrates our pipeline. The first stage involves learning a scene-specific degradation mapping (SDM) network to obtain low-resolution results from high-resolution input by performing internal learning on a coarsely trained NeRF (i.e. coarse NeRF). In the second stage, we learn a fine super-resolution NeRF (i.e. fine NeRF) by using the low-resolution scene data (i.e. ground truth), where the learned SDM model is used to guide the training of the fine NeRF by conducting inverse rendering. Next, we introduce the details in different subsections. In Section \ref{4.1}, we present how to obtain the SDM model by using the internal learning method. In Section \ref{4.2}, we present how to obtain the fine NeRF model by using our super-resolution training method. In Section \ref{4.3}, we combine radiance fields from multiple time steps in a single training process to alleviate the variance in super-resolution scene estimation.

\subsection{Internal Learning of Radiance Fields}
\label{4.1}
As shown in Figure \ref{pipeline} (a), we establish a SDM model $\mathcal{M}$ based on deep convolutional neural network (CNN) to downsample a rendered view $\hat{I}_{LR}$ and produce the following low-resolution result $\hat{I}_{LLR}$:
\begin{equation}
\hat{I}_{LLR}=\mathcal{M}\left(G_{LR} , \hat{I}_{LR}\right)
\end{equation}
where $G_{LR}$ denotes the corresponding \textbf{gradient view} of $\hat{I}_{LR}$.
To train our \textbf{SDM network}, we introduce a single-scene internal learning method in the radiance fields of NeRF. First, we pretrain a \textbf{coarse NeRF} $C^{*}_{coarse}$ by using sparse ray sampling and then employ it to render a novel view $\hat{I}_{LR}$. Second, we use the down-sampled result $I_{LLR}$ (H/s x W/s) of the ground truth image $I_{LR}$ (H x W) to supervise the internal learning, where $s$ denotes the scale factor. The goal is to optimize the following loss function by using mean square error (MSE):
\begin{equation}
\mathcal{L}_{MSE}=\left\|\hat{I}_{LLR}-I_{LLR}\right\|_2^2.
\end{equation}
Note that the above internal learning method can enable us to (a) avoid the consumption of additional scene data, and (b) simulate a realistic and complex down-sampling effect.


\textbf{SDM network.} We propose to exploit a lightweight CNN to realize our SDM model by performing layer-wise down-sample. To avoid bringing convolution to a higher dimension, we use Pixel Adaptive Convolution \cite{PAC} (PAC) as the down-sample layer, which generates adaptive convolution weights for each pixel based on the provided guidance information (i.e. $G_{LR}$) to fit complex mapping relationships:
\vspace{-0.1cm}
\begin{equation}
v_i^{\prime}=\sum_{j\in\Omega(i)}K\left(f_i,f_j\right)W\left[p_i-p_j\right]v_j+b
\vspace{-0.3cm}
\end{equation}
where $\Omega(i)$ denotes the neighbors of pixel $i$, $K$: $\mathbb{R}^{c^{\prime}\times c\times s\times s}$ denotes the kernel function, $W$ denotes the weight and $b$ denotes the bias. Another advantage of using PAC lies in that it can effectively establish and transfer the mapping relationship across scales, which can enable us to train the SDM model by using low-resolution data and generalize it to process the high-resolution data. As a result, our internal learning method can effectively model the internal recurrence of information across the different scales. Different from most existing works \cite{NVSR,NeRF-SR} that lack in-depth mining of the information hidden in a single scene, SDM takes the scene structure and surrounding pixel information into consideration to model the transformation between different data scales.

\textbf{Coarse NeRF.} To facilitate the down-sampling process, we initially train a coarse NeRF by using patches of low-resolution rays and low-resolution ground truth images $I_{LR}$ (HxW). A novel view can be rendered by using:
\vspace{-0.1cm}
\begin{equation}
\hat{I}_{LR}=C_{\text {coarse}}^*\left(P_{LR}\right)
\vspace{-0.5cm}
\end{equation}
where $P_{LR}$ denotes the low-resolution patches of rays.

\textbf{Gradient View} $G$ is used to regularize the possible solutions of SDM model. To obtain $G$, We first utilize the Sobel operator to extract the horizontal and vertical gradients of the rendered low-resolution image $\hat{I}_{LR}$:
\vspace{-0.3cm}
\begin{equation}
D_u=\left[\begin{array}{lll}
-1 & 0 & 1 \\
-2 & 0 & 2 \\
-1 & 0 & 1
\end{array}\right] * \hat{I}_{LR}
\end{equation}
\vspace{-0.3cm}
\begin{equation}
D_v=\left[\begin{array}{ccc}
-1 & -2 & -1 \\
0 & 0 & 0 \\
1 & 1 & 2
\end{array}\right] * \hat{I}_{LR}
\end{equation}
Then, we calculate the following gradient magnitude as $G_{LR}$ to summarize the edge information in $\hat{I}_{LR}$:
\begin{equation}
G_{LR}=\sqrt{D_u{ }^2+D_v{ }^2}
\end{equation}

\subsection{SDM-Guided Super-Resolution Training}
\label{4.2}

Initially, the coarse NeRF is trained by using an available low-resolution ground truth dataset of HxW, which can render high-fidelity novel views at the same resolution. However, when performing super-resolution rendering at a higher scale factor of $s$, it may inevitably confront the blurring effect because there exist many sampling points (or sampling gaps) whose values are usually undefined due to the lack of detailed information at higher resolutions. The distortion would become more severe for larger $s$.

To address the problem, we propose a SDM-guided super-resolution training method to obtain a fine NeRF, $C_{fine}$, by sampling a grid of $s^2$ rays within each pixel (i.e. super-sampling) and filling in the high-frequency details that are undefined in the low-resolution NeRF. The main difficulty lies in the sub-pixel supervision of the super-resolution (i.e. sH×sW) rendering results by using a low-resolution (i.e. H×W) ground truth view. As shown in Figure \ref{pipeline} (b), we adopt an inverse rendering strategy by generalizing the well-learned SDM model to assist the training of the fine NeRF, where the resolution of the output of the fine NeRF is sH×sW, the input and output resolutions of SDM model are sH×sW and H×W, respectively.

\textbf{Inverse Rendering.} Instead of using the single ray-based optimization method, we advocate using the patch-wise optimization to capture the correlated ray information within the spatial neighborhood, which would favor the internal learning of the scene information. A novel super-resolution view $\hat{I}_{HR}$ can be rendered by using the patches of high-resolution rays $P_{HR}$:
\begin{equation}
\hat{I}_{HR}=C_{fine}\left(P_{HR}\right)
\end{equation}
where the fine NeRF $C_{fine}$ is updated in the training process. Then, the obtained high-resolution patch $\hat{I}_{HR}$ is mapped to its corresponding low-resolution version $\hat{I}_{LHR}$ by performing inverse rendering using our pretrained SDM model $\mathcal{M}^*$:
\begin{equation}
\hat{I}_{LHR}=\mathcal{M}^*\left(G_{HR}, \hat{I}_{HR}\right)
\end{equation}
where $G_{HR}$ denotes the gradient view of $\hat{I}_{HR}$. The resolution of $\hat{I}_{LHR}$ is HxW, which is the same as that of the ground truth training data.

\textbf{Objective Function.} The overall objective of super-resolution training is to optimize the following loss function:
\begin{equation}
\mathcal{L}=\mathcal{L}_{MSE}+\lambda \mathcal{L}_{perc}
\end{equation}
where $\lambda$ denotes the hyper-parameter. $\mathcal{L}_{MSE}$ is the MSE loss used to optimize the parameters of the fine NeRF:
\begin{equation}
\mathcal{L}_{MSE}=\left\|\hat{I}_{LHR}-I_{LR}\right\|_2^2.
\end{equation}
$\mathcal{L}_{perc}$ is the patch-based perceptual loss used to improve high-frequency details by estimating the similarity between predicted patches $\hat{I}_{LHR}$ and ground-truth $I_{LR}$ in the feature space via a pretrained 19-layer VGG network $\varphi$:
\begin{equation}
\mathcal{L}_{perc}=\left\|\varphi(\hat{I}_{LHR})-\varphi(I_{L R})\right\|_2^2.
\end{equation}

\textbf{Coarse-to-Fine Optimization.} To effectively train our model, we present a coarse-to-fine strategy to speed up the convergence. We use the first $N_{1}$ epochs to train the coarse NeRF. Then, we use $N_{2}$ epochs to train the fine NeRF. The employed pretrained SDM model is optimized separately before training the fine NeRF by using $N_{3}$ epochs. Note that by integrating the information from multiple views during the training process, our inverse rendering strategy can bridge more information to fill in the undefined gaps in super-resolution rendering, which can help to handle the complex rendering challenges.


\subsection{Temporal Ensemble}
\label{4.3}
The super-resolution training of 3D NeRF is analogous to that of 2D images. Since it is an inverse problem, there is more than one solution for the high-quality NeRF for the corresponding low-resolution view. In fact, from the perspective of ill-posed problems, the constraints on a gradient are equivalent to the regularization, allowing the gradient descent method to maintain a relatively ideal optimization direction. Although regularization can narrow the solution space, local degradation still exists, which may affect the robustness of the results.

\begin{figure}[htb]
\centering
\includegraphics[width=0.5\textwidth]{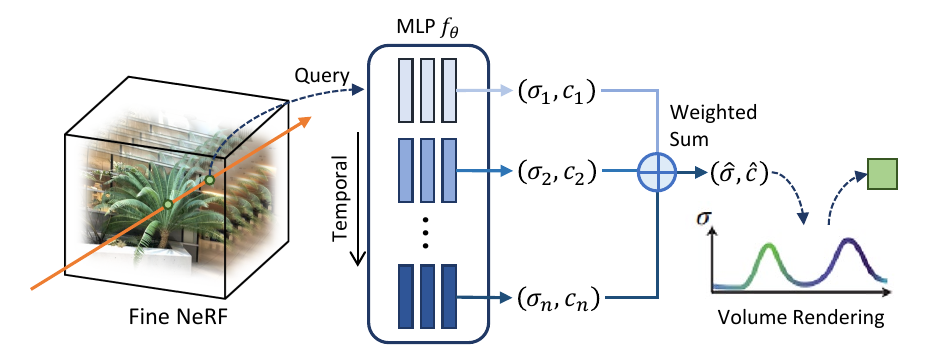}
\vspace{-0.5cm}
\caption{Illustration of the temporal ensemble strategy. The radiance fields at multiple time steps are ensembled with respect to color and density.}
\label{temporal ensemble}
\vspace{-0.6cm}
\end{figure}

To deal with the problem, we further propose a temporal ensemble strategy to narrow the solution space and obtain a more smooth estimation of density and color corresponding to 5D coordinates. Figure \ref{temporal ensemble} illustrates the records of the radiance fields at multiple time steps. The estimated density and color information are averaged by using
\begin{align}
\hat{\sigma}, \hat{c} = \frac{1}{N} \sum_{i}^{N} \sigma_i, \frac{1}{N} \sum_{i}^{N} c_i
\end{align}
where $N$ denotes the number of radiance fields to be ensembled, $\sigma_i$ and $c_i$ denote the estimated volume density and view-dependent color. We need to point out that our temporal ensemble strategy can compensate for the errors of the estimated super-resolution scene, effectively alleviating the inescapable local degradation issue associated with super-resolution training.

\begin{figure*}[htb]
\centering
\scriptsize
\begin{tabular}{cccccc}

\makecell[c]{
\includegraphics[trim={0px 0px 0px 0px}, clip, width=\resultsfigwidth]{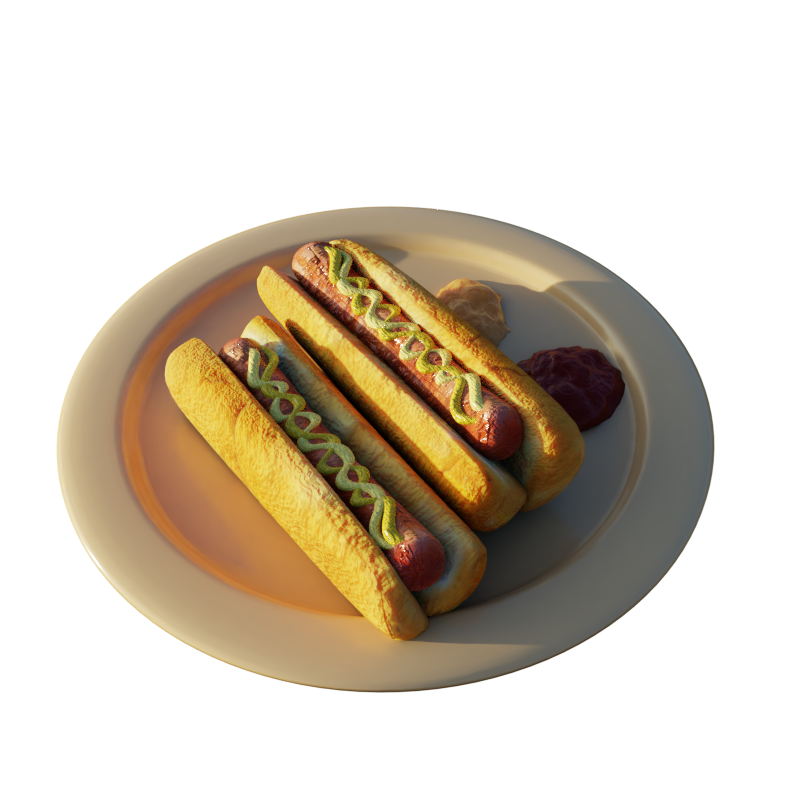}
\put(-70,60){{\color{red}{\huge$\square$}}}
\put(-56,58){{\color{green}{\huge$\square$}}} \\
}
&
\crophotdog{photo/visual_blender/hotdog_gt.png}{Ground Truth} &
\crophotdog{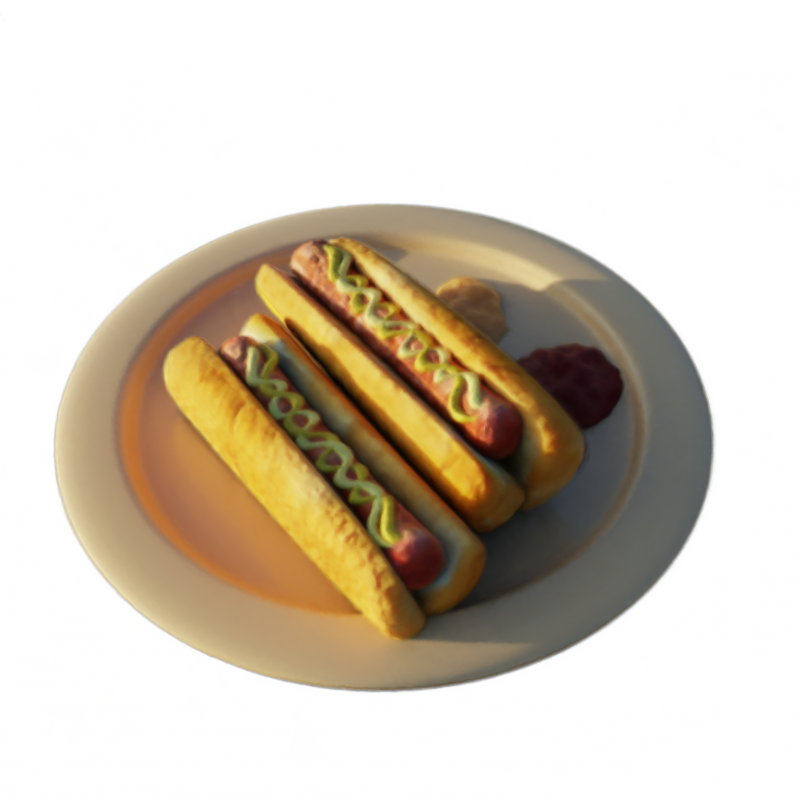}{NeRF} &
\crophotdog{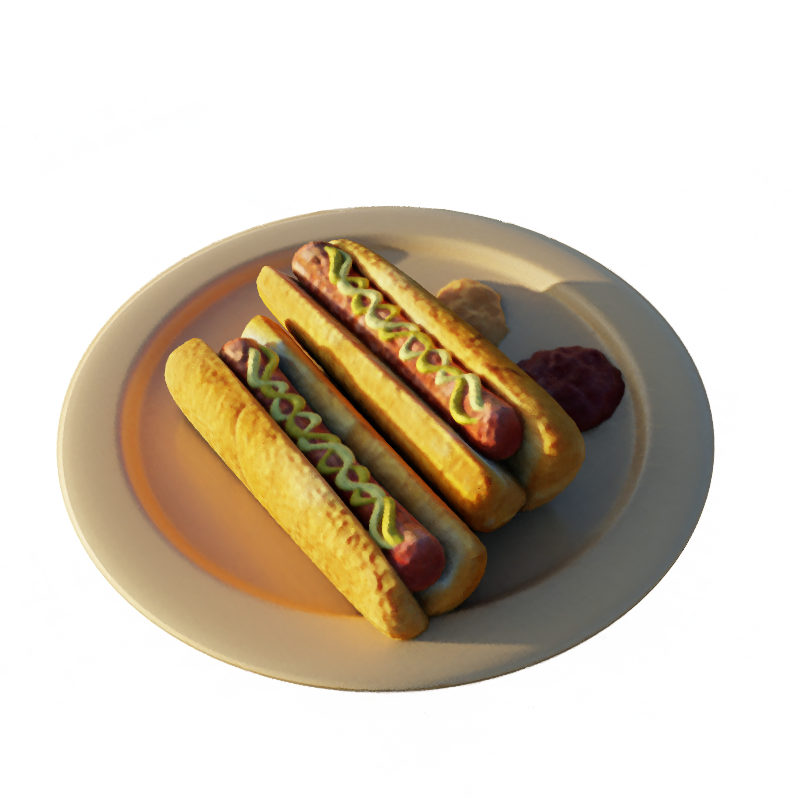}{NeRF-SR} &
\crophotdog{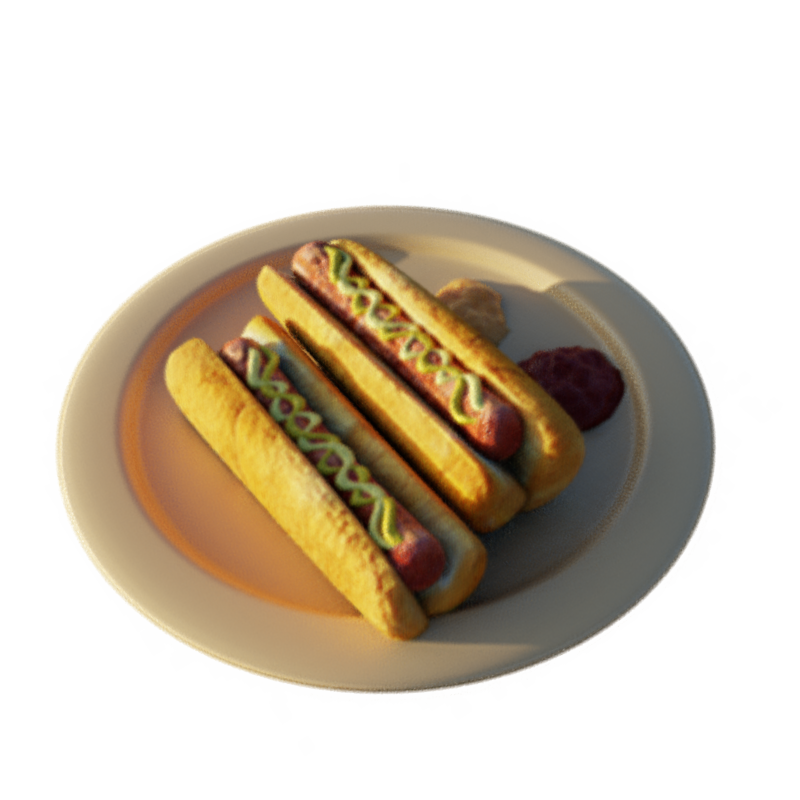}{TensoRF} &
\crophotdog{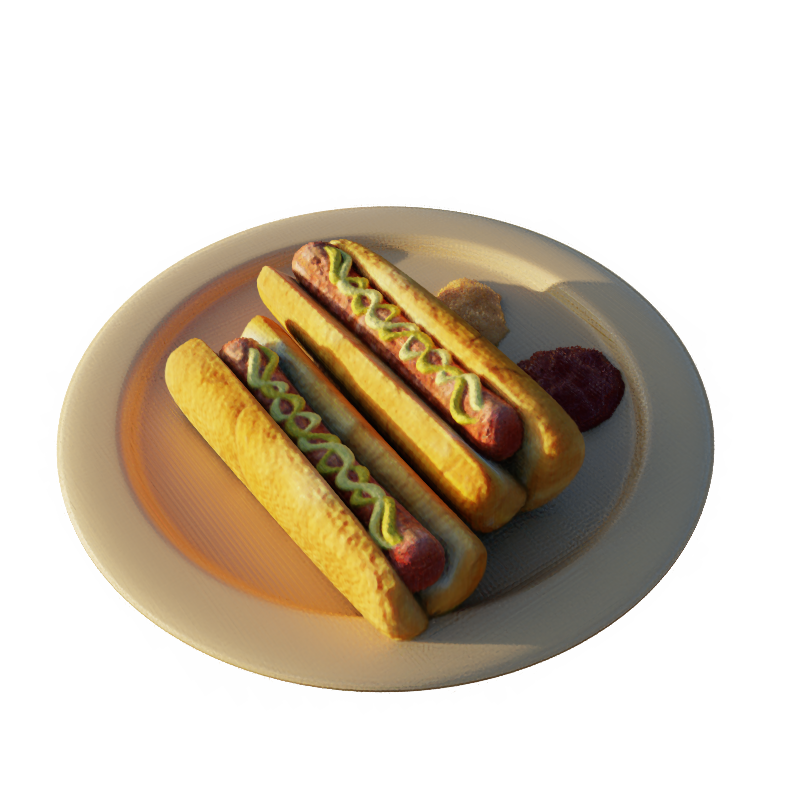}{Ours} \\

\makecell[c]{
\includegraphics[trim={0px 0px 0px 0px}, clip, width=\resultsfigwidth]{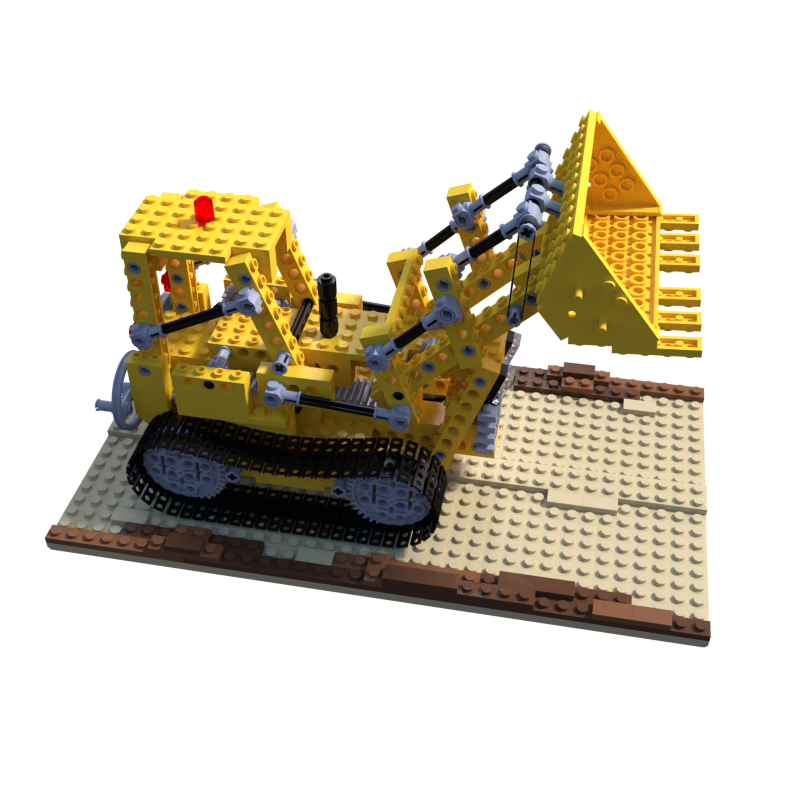}
    \put(-62,40){{\color{red}{\huge$\square$}}}
    \put(-45,60){{\color{green}{\huge$\square$}}} \\
}
&
\croplego{photo/visual_blender/lego_gt.png}{Ground Truth} &
\croplego{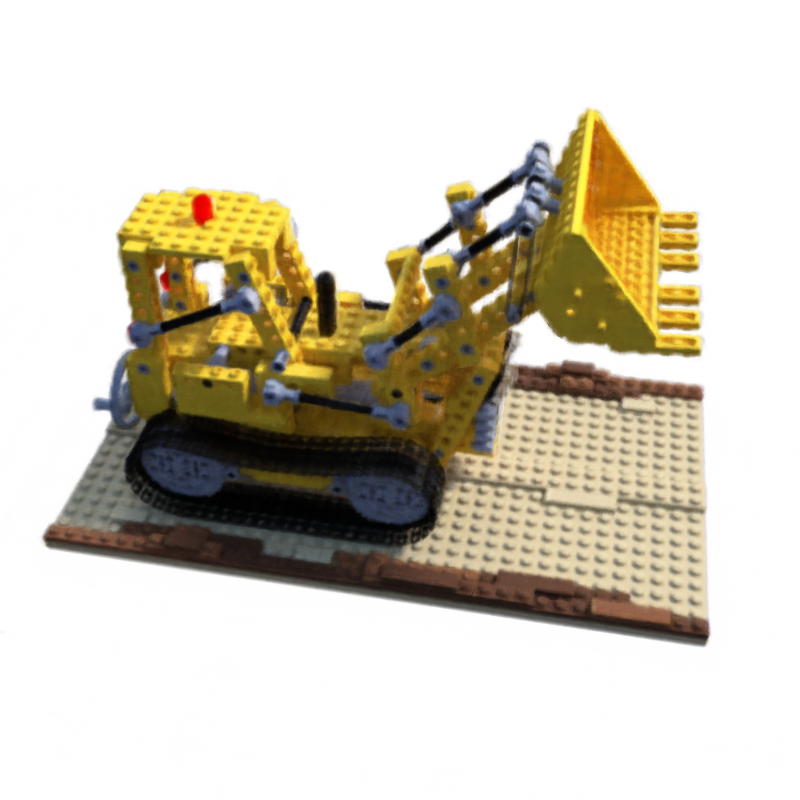}{NeRF} &
\croplego{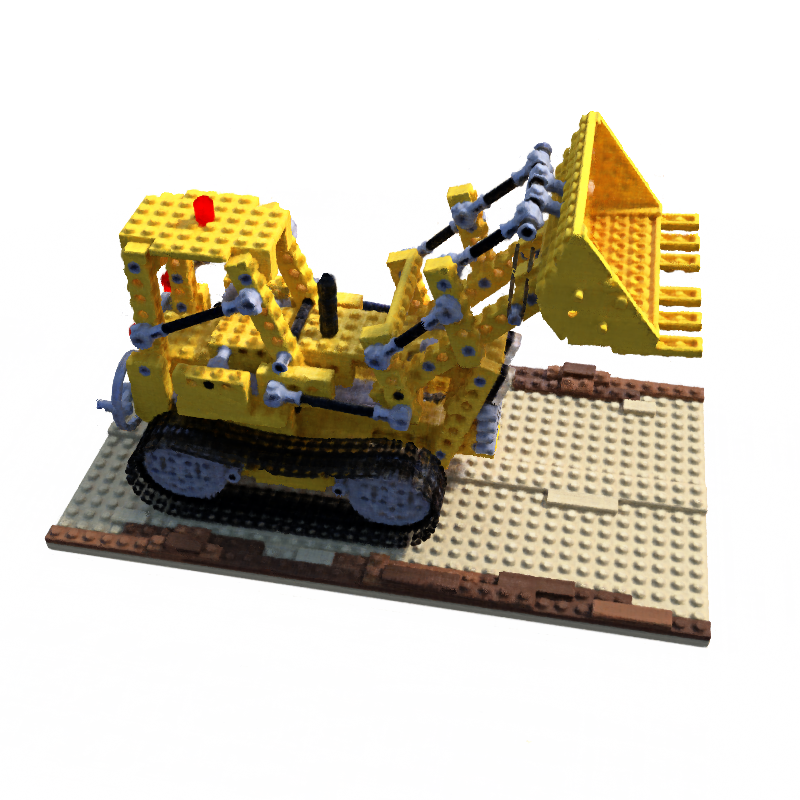}{NeRF-SR} &
\croplego{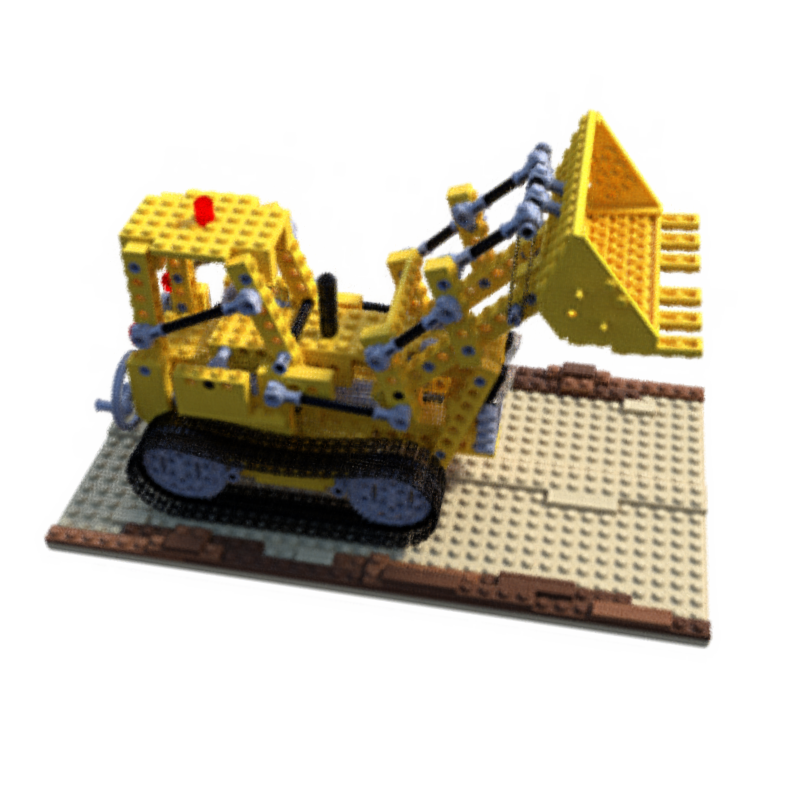}{TensoRF} &
\croplego{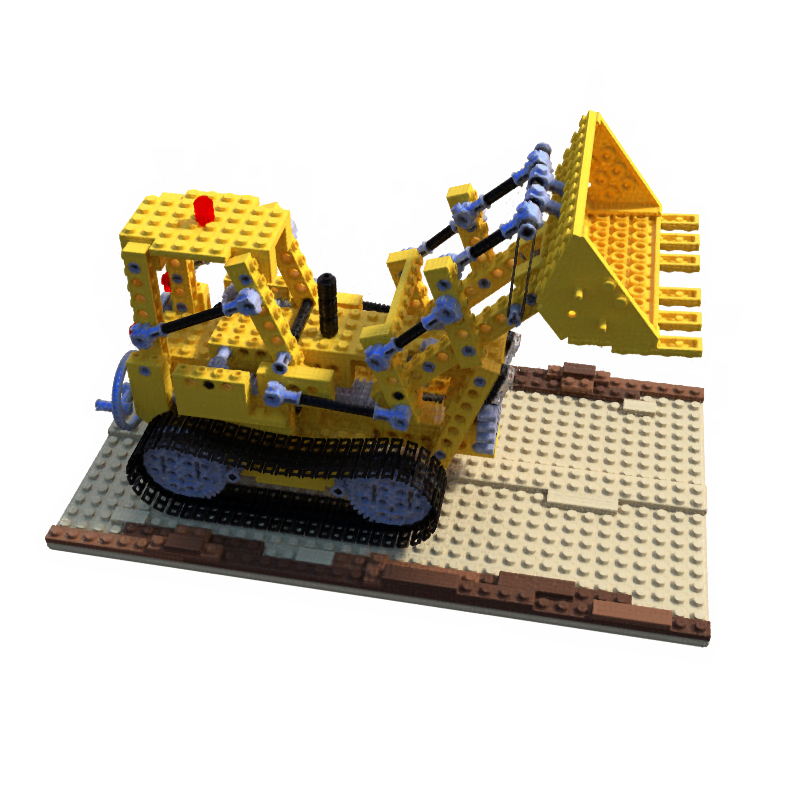}{Ours} \\

\makecell[c]{
	\includegraphics[trim={0px 0px 0px 0px}, clip, width=\resultsfigwidth]{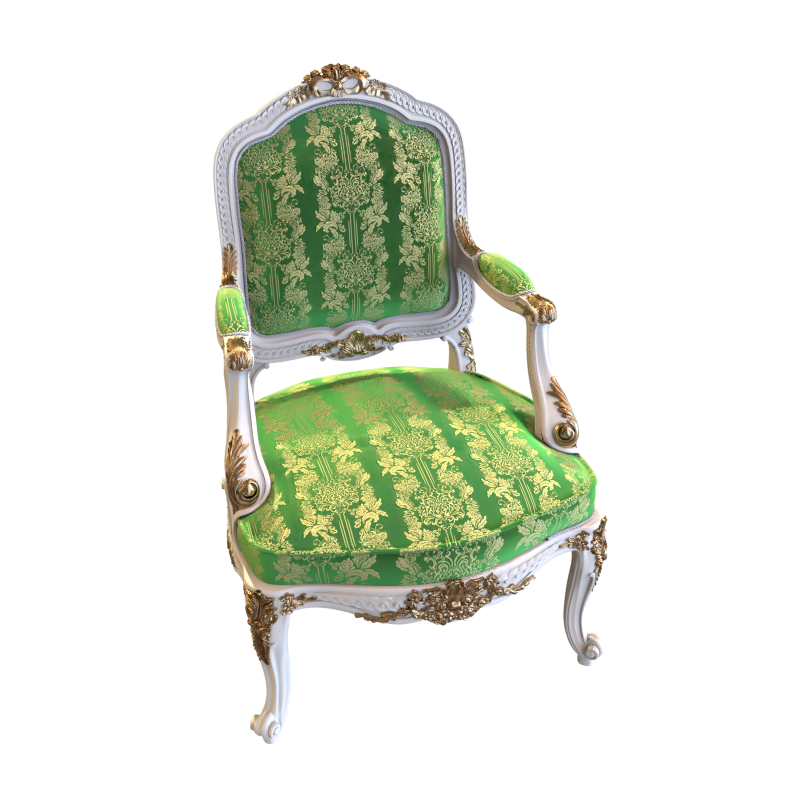}
	\put(-50,40){{\color{red}{\huge$\square$}}}
 \put(-65,50){{\color{green}{\huge$\square$}}} \\
}
&
\cropchair{photo/visual_blender/chair_gt.png}{Ground Truth} &
\cropchair{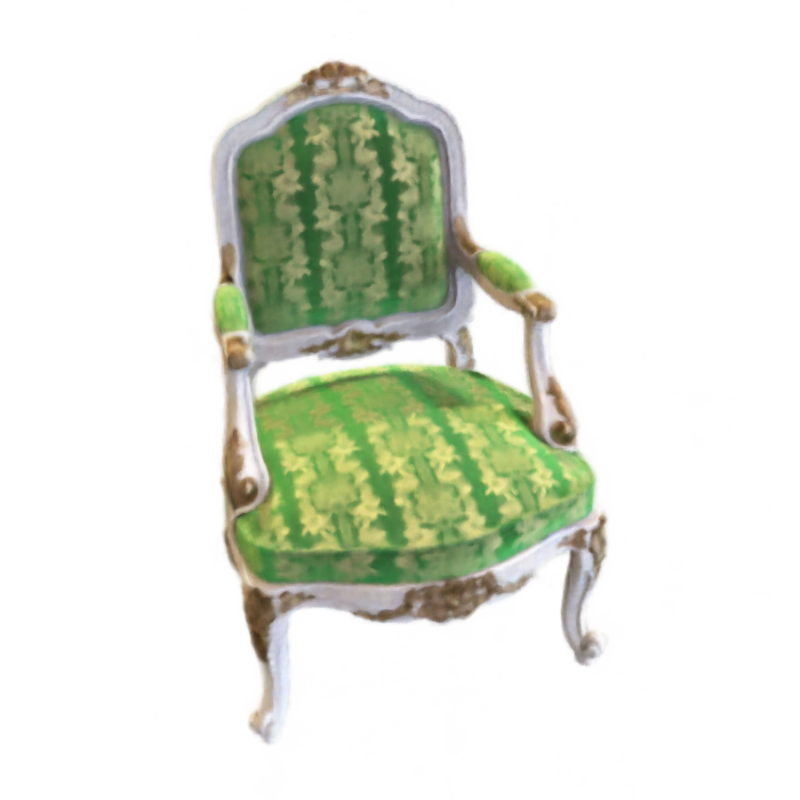}{NeRF} &
\cropchair{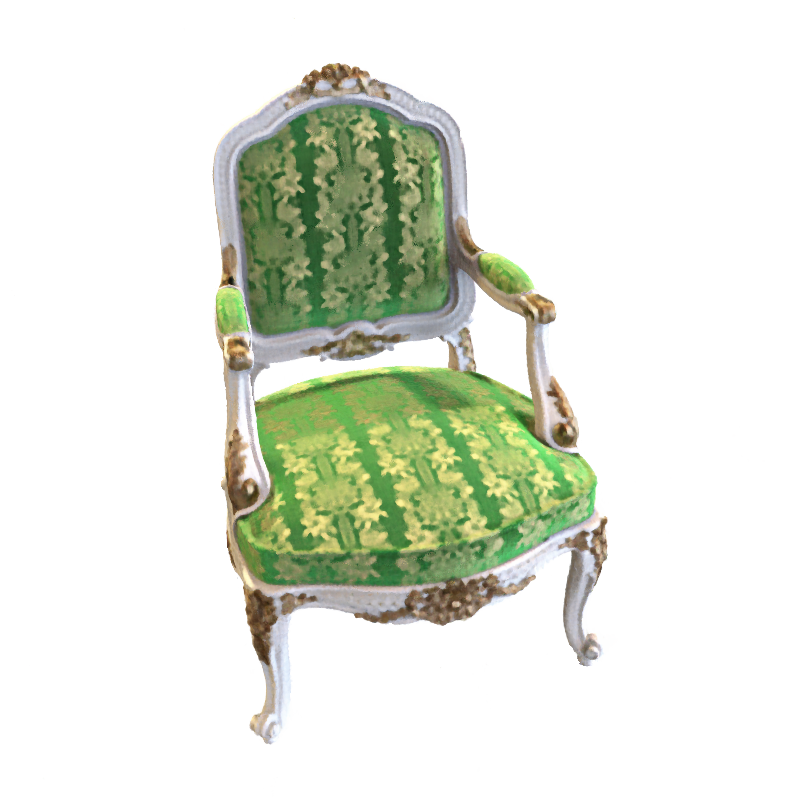}{NeRF-SR} &
\cropchair{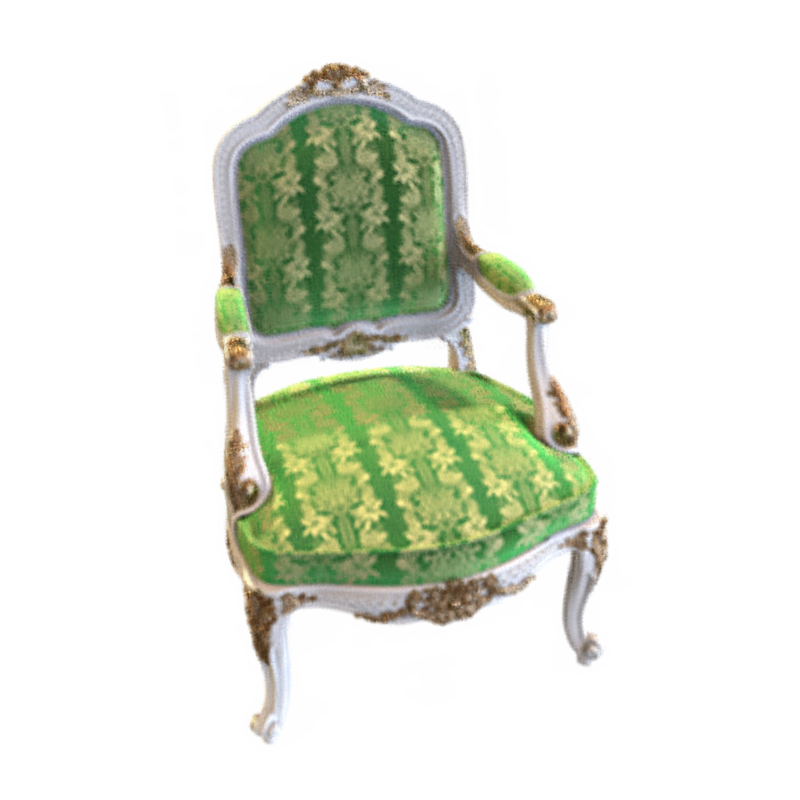}{TensoRF} &
\cropchair{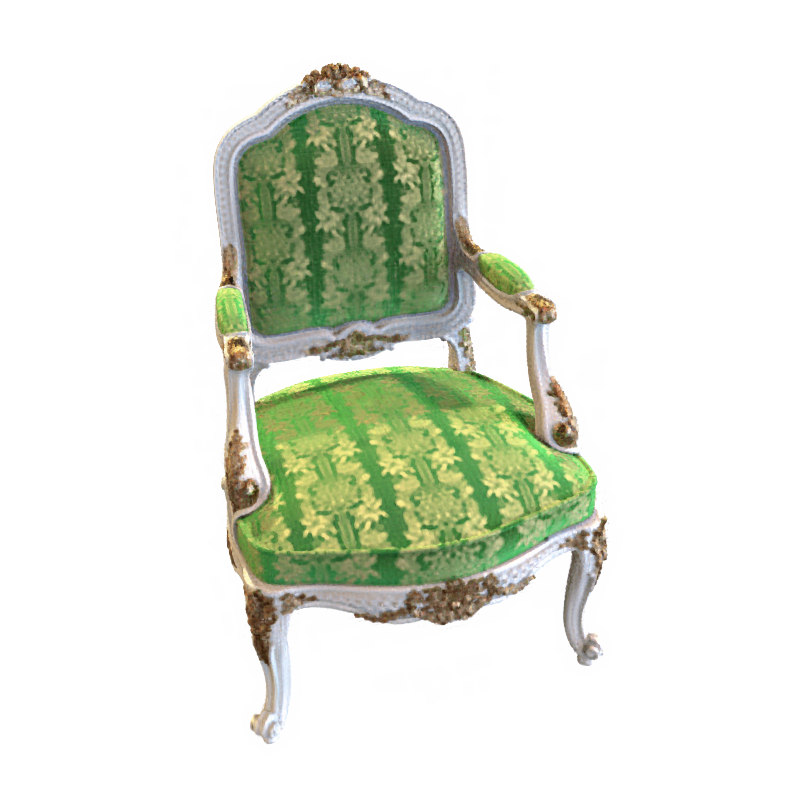}{Ours} \\

\makecell[c]{
	\includegraphics[trim={0px 0px 0px 0px}, clip, width=\resultsfigwidth]{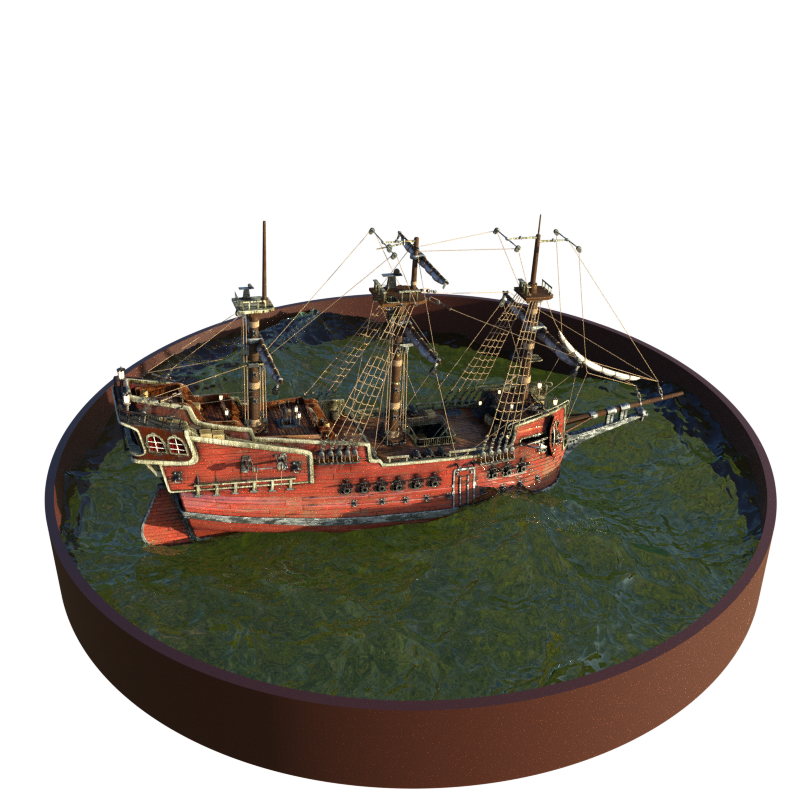}
        \put(-62,55){{\color{red}{\huge$\square$}}}
	\put(-90,45){{\color{green}{\huge$\square$}}} \\
}
&
\cropship{photo/visual_blender/ship_gt.png}{Ground Truth} &
\cropship{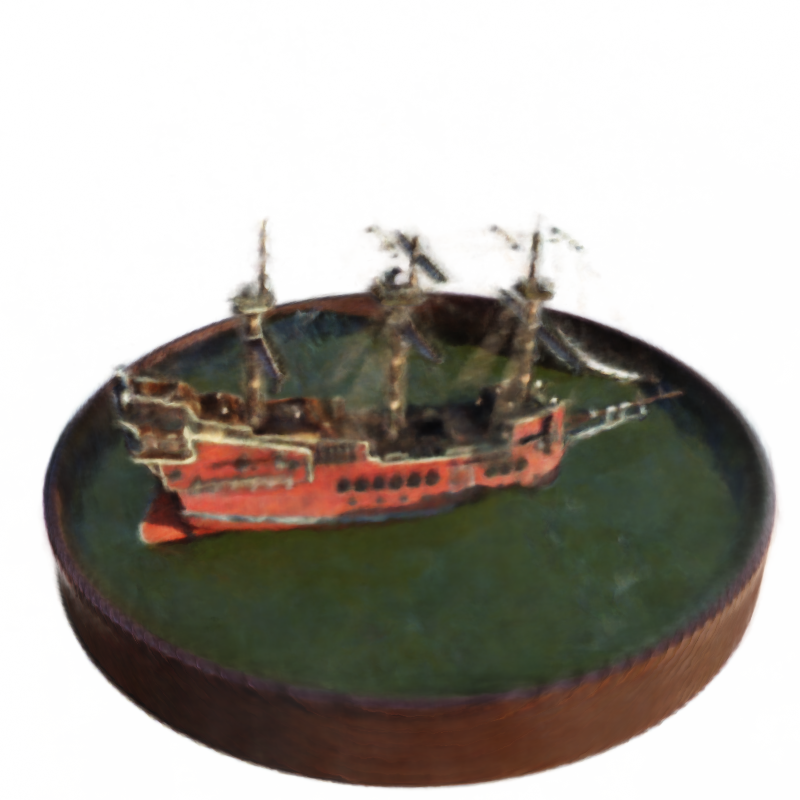}{NeRF} &
\cropship{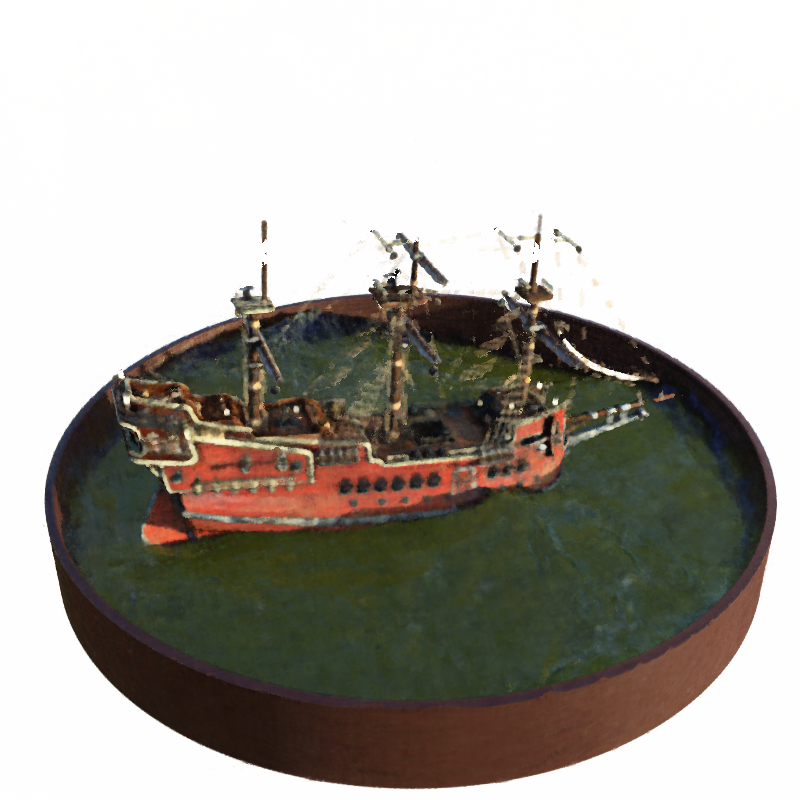}{NeRF-SR} &
\cropship{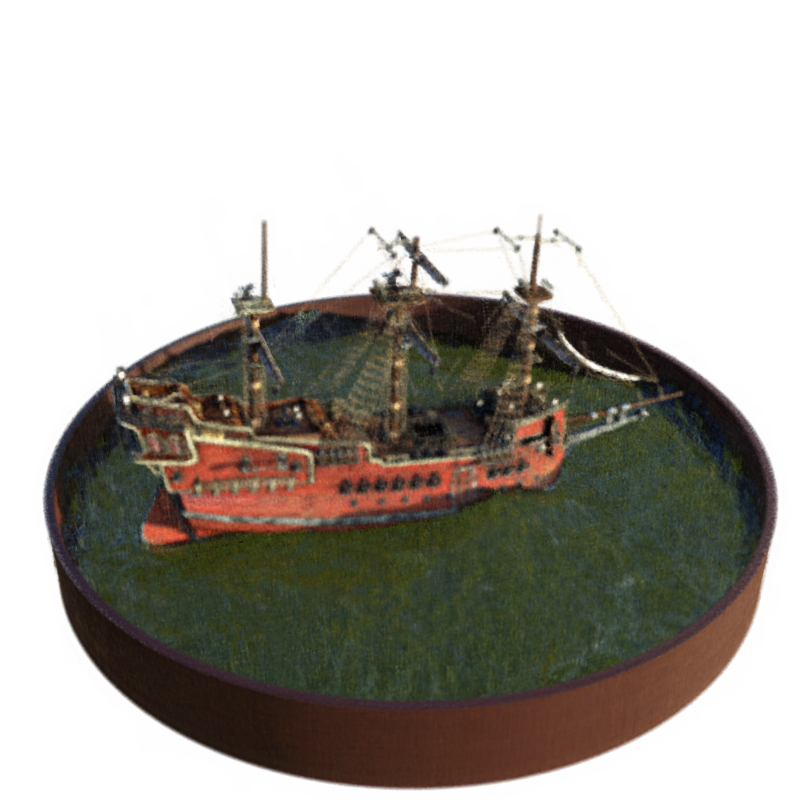}{TensoRF} &
\cropship{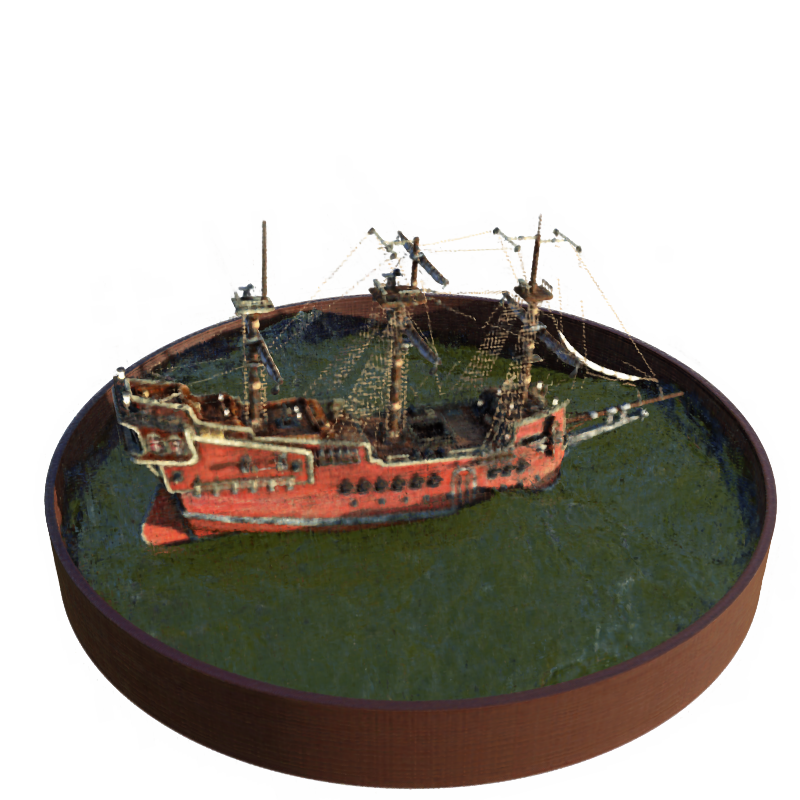}{Ours} \\


\end{tabular}
\vspace{-0.1cm}
\caption{Comparison of the synthesized super-resolution views (x4) of different methods on the Blender dataset. For each of the views in the first column, the results in the first line show the larger version of the red box region, and the results in the second line show the larger version of the green box region. Our results show clearer details than NeRF, NeRF-SR, and TensoRF.}
\label{visual_blender}
\vspace{-0.5cm}
\end{figure*}

\section{Experiment}
In this section, we conduct quantitative and qualitative studies to show the performance of the proposed method.

\subsection{Datasets}

\textbf{Blender Dataset} is a popular public dataset that contains eight synthetic objects. Each scene is captured from virtual cameras positioned in a hemisphere layout, focusing inwards. The dataset includes 100 training images per scene and reserves 200 images for testing. All images are in a high-resolution format of 800×800.

\textbf{LLFF Dataset} is a popular public dataset that comprises eight real-world scenes, primarily composed of forward-facing images. These scenes are captured using 20 to 62 images each. For testing purposes, 1/8 of these images are used as the test set. The resolution of all images in this dataset is 1008×756.

\subsection{Implementation Details}
\textbf{Basic Settings.} TensoRF-VM-192 is used to implement our NeRF model, and the configurations, such as tensor resolution and number, remain unchanged. We use the Adam optimizer with initial learning rates of 0.02 for tensor factors and 0.001 for the MLP decoder. All our experiments are done on a single RTX 3090 GPU.

\textbf{Patch-wise Ray Sampling.} To address the issue of empty space in the blender dataset, we construct a mask for each patch to speed up training using the density estimation of the coarse NeRF. In the Super-Resolution Training stage, we set patch size to 16x16 (x2) and 32x32 (x4), and the corresponding batch sizes are set to 32 and 8.

\textbf{Coarse NeRF.} The configuration of training the coarse NeRF is $N_{1}=$5,000 for the blender dataset, and $N_{1}=$10,000 for the LLFF dataset, and the batch size is set to 4096 for the randomly sampled ray. 

\textbf{Fine NeRF.} The configuration of training the fine NeRF is $\lambda$=0.03, $N_{2}$=25,000 for the blender dataset and $N_{2}$=20,000 for the LLFF dataset, and the batch size is set to 8192. 

\textbf{SDM Model.} The configuration of training the SDM model is $N_{3}$=10,000, patch size 16 (x2), 32 (x4) and the
corresponding batch sizes are set to 32 and 8.

\begin{figure*}[htbp]
\centering
\scriptsize
\begin{tabular}{cccccc}

\makecell[c]{
\includegraphics[trim={0px 0px 0px 0px}, clip, width=\resultsfigwidth]{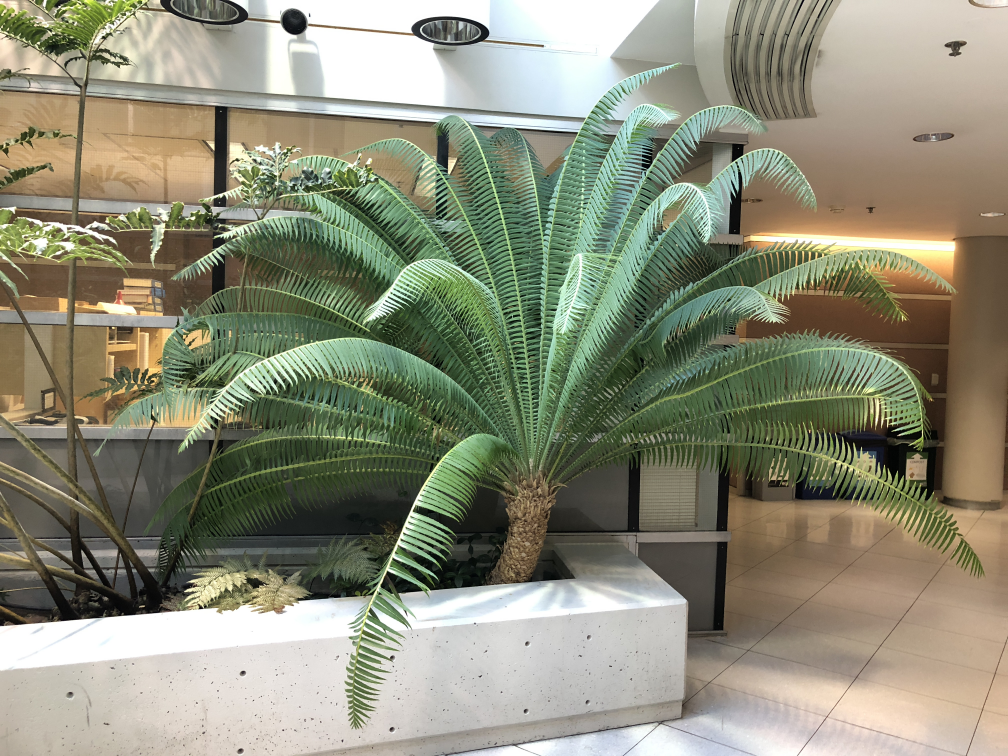}
\put(-25,30){{\color{red}{\huge$\square$}}}
\put(-43,53){{\color{green}{\huge$\square$}}} \\
}
&
\cropfern{photo/visual_llff/fern_gt.png}{Ground Truth} &
\cropfern{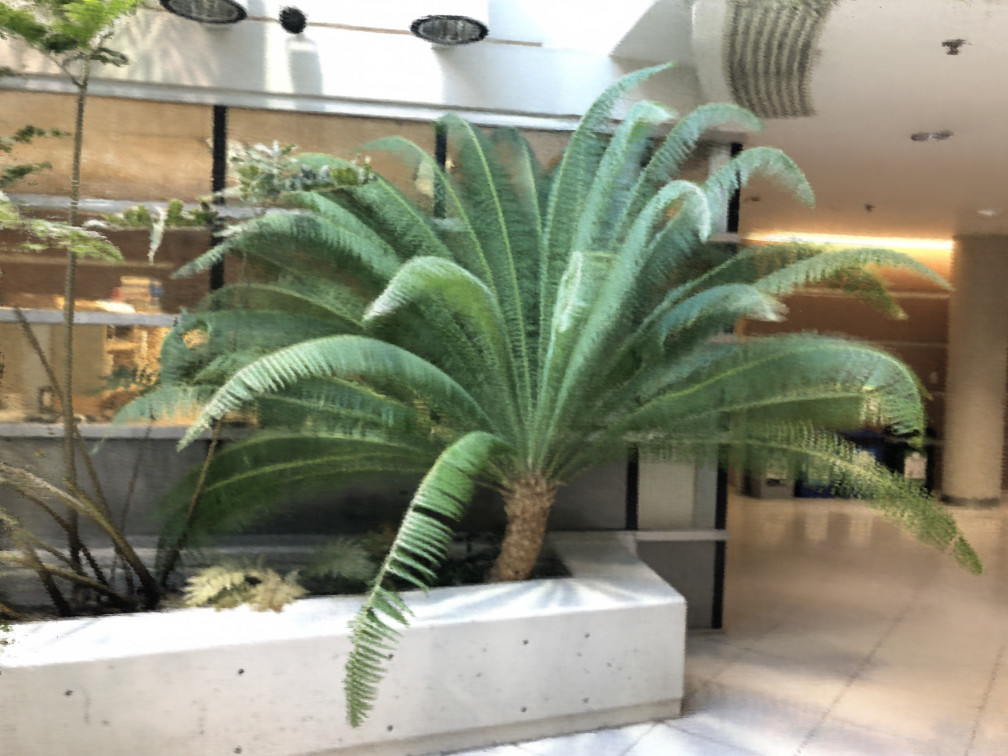}{NeRF} &
\cropfern{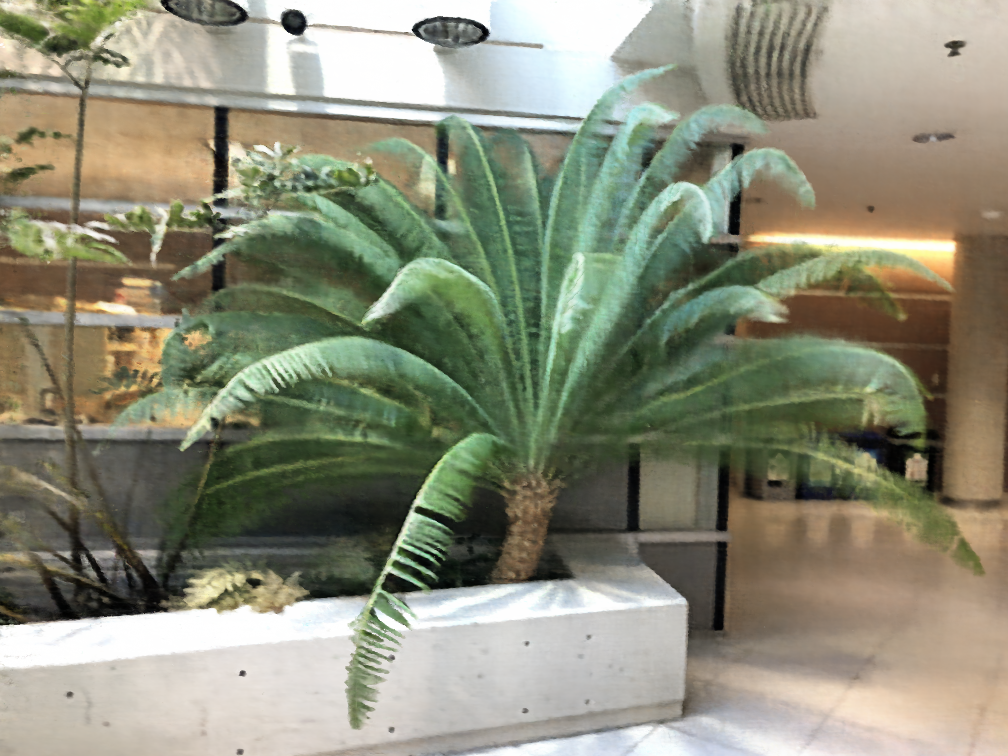}{NeRF-SR} &
\cropfern{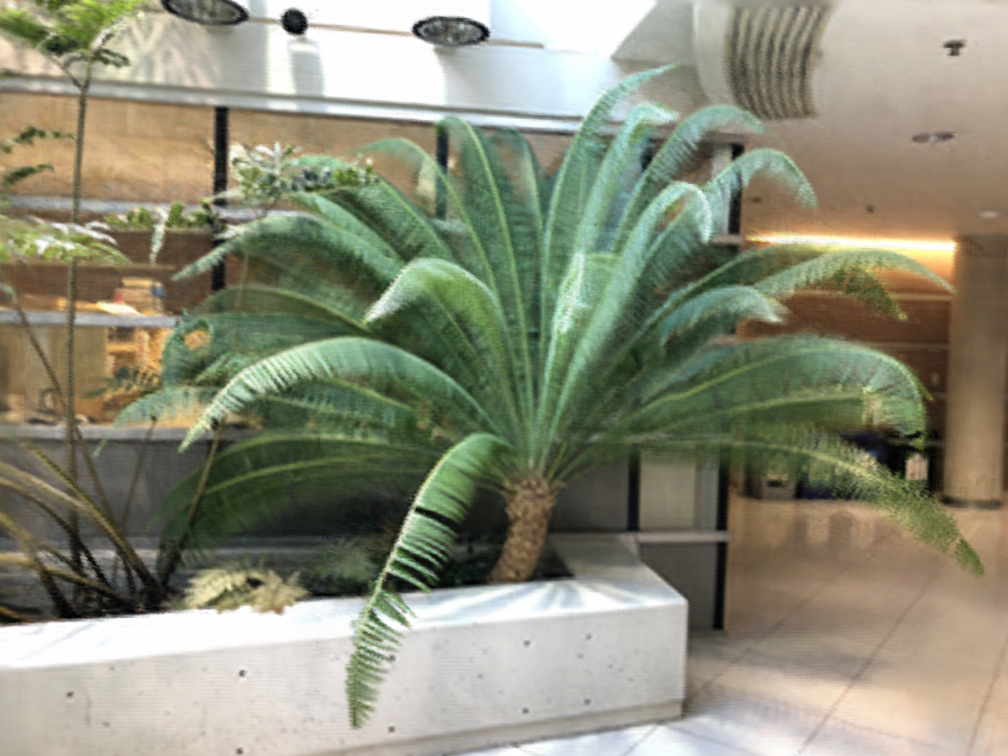}{TensoRF} &
\cropfern{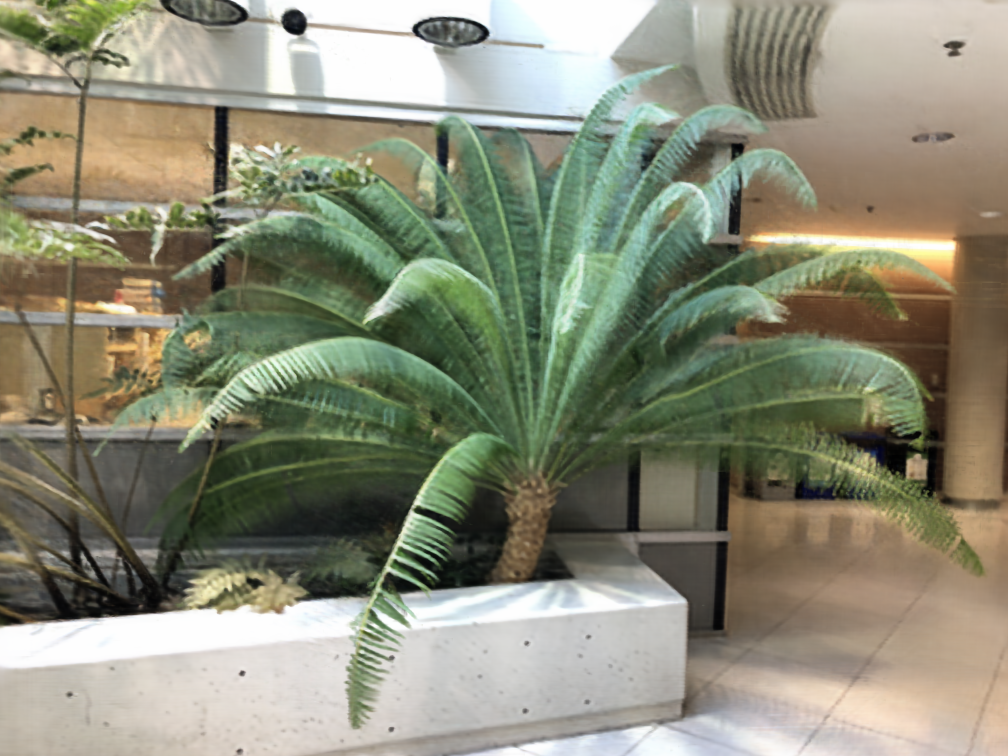}{Ours} \\

\makecell[c]{
\includegraphics[trim={0px 0px 0px 0px}, clip, width=\resultsfigwidth]{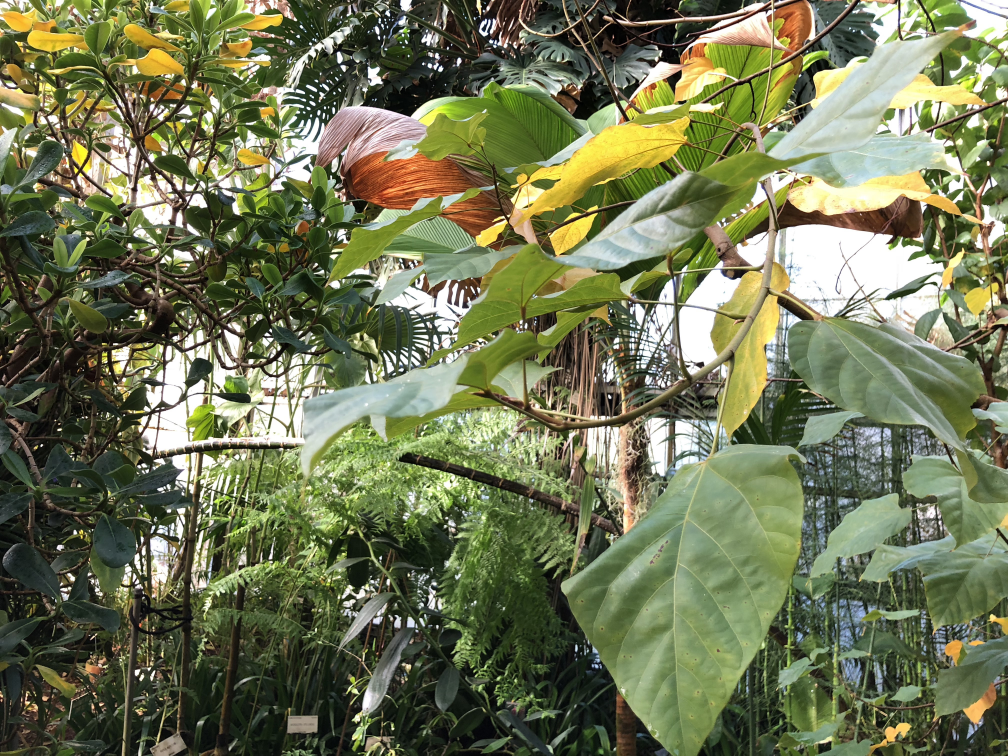}
\put(-43,15){{\color{red}{\huge$\square$}}}
\put(-43,53){{\color{green}{\huge$\square$}}} \\
}
&
\cropleaves{photo/visual_llff/leaves_gt.png}{Ground Truth} &
\cropleaves{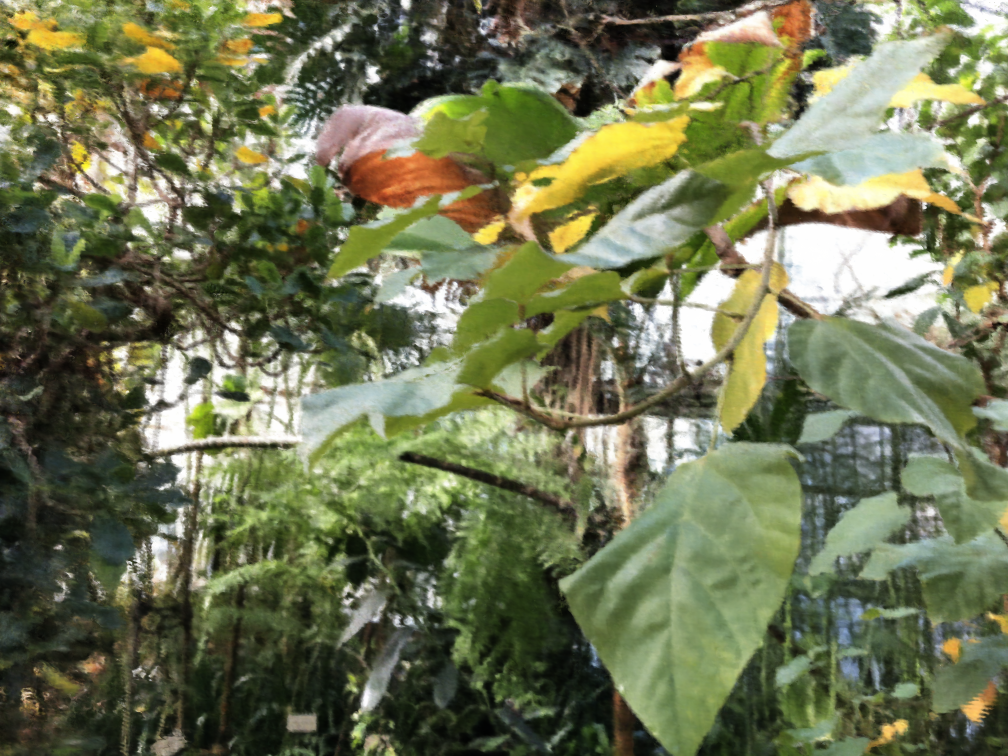}{NeRF} &
\cropleaves{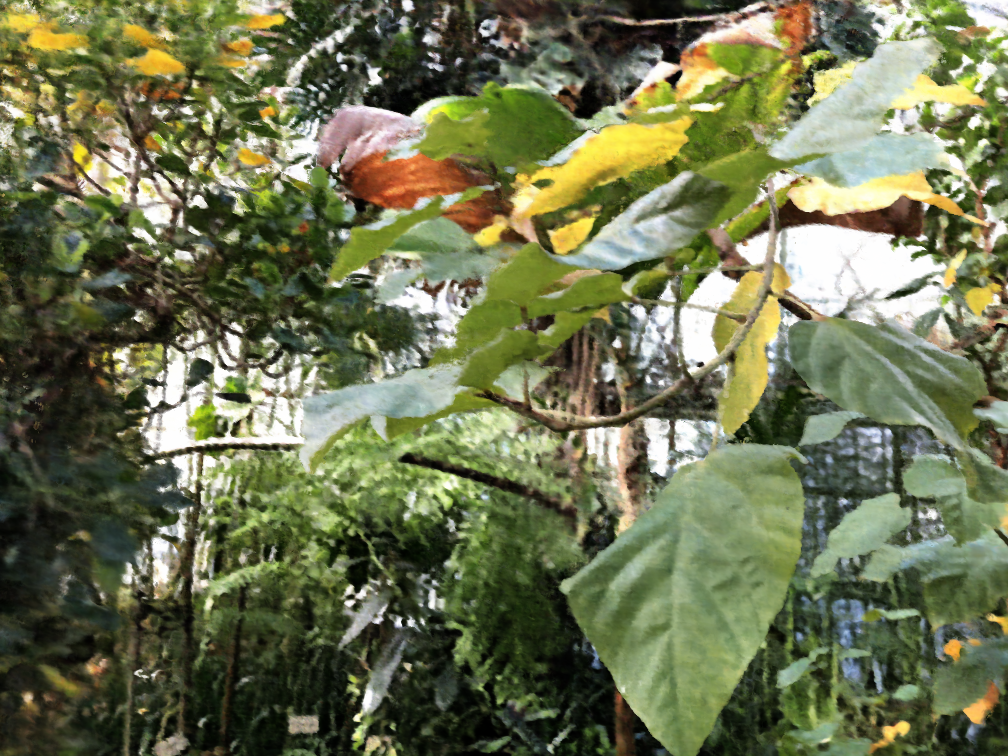}{NeRF-SR} &
\cropleaves{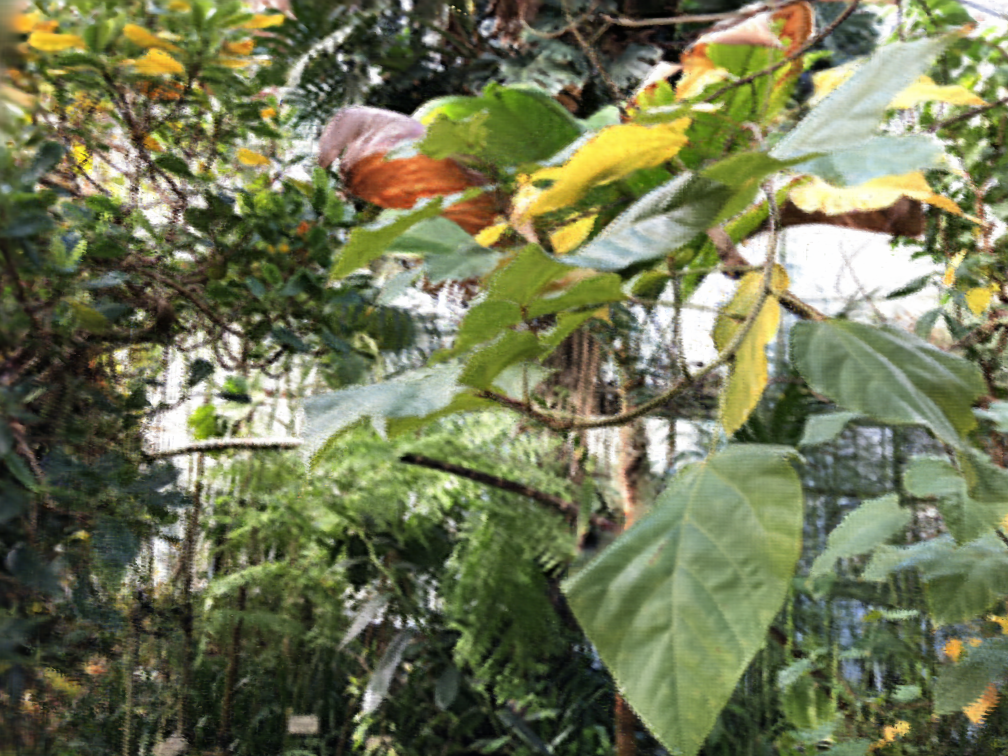}{TensoRF} &
\cropleaves{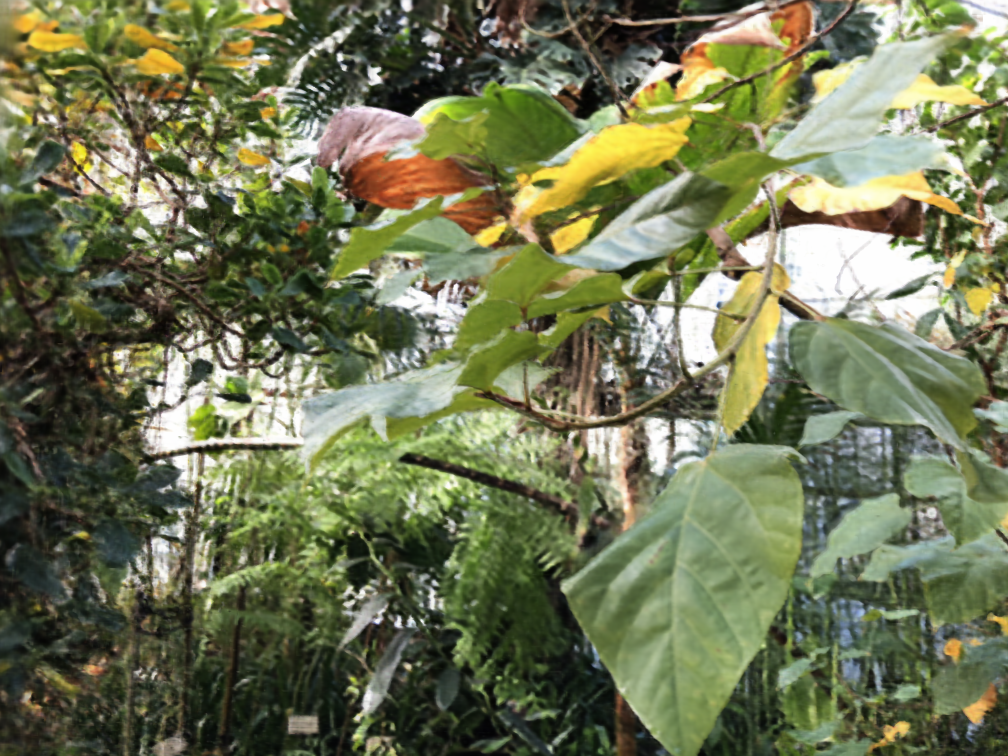}{Ours} \\

\makecell[c]{
\includegraphics[trim={0px 0px 0px 0px}, clip, width=\resultsfigwidth]{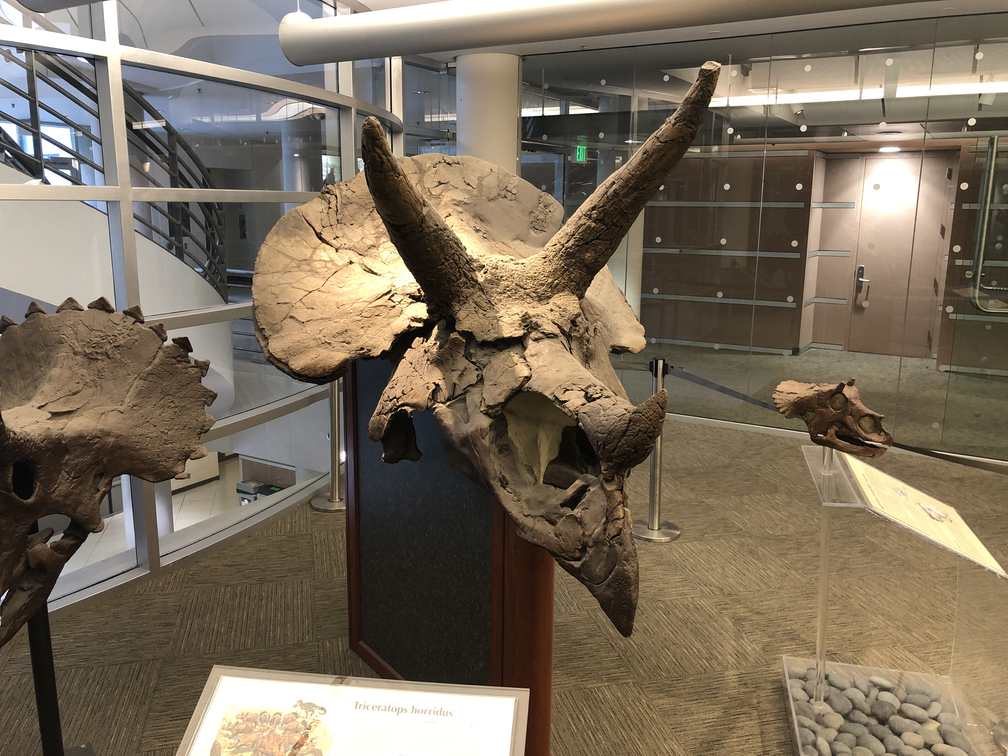}
\put(-28,27){{\color{red}{\huge$\square$}}}
\put(-93,53){{\color{green}{\huge$\square$}}} \\
}
&
\crophorn{photo/visual_llff/horn_gt.png}{Ground Truth} &
\crophorn{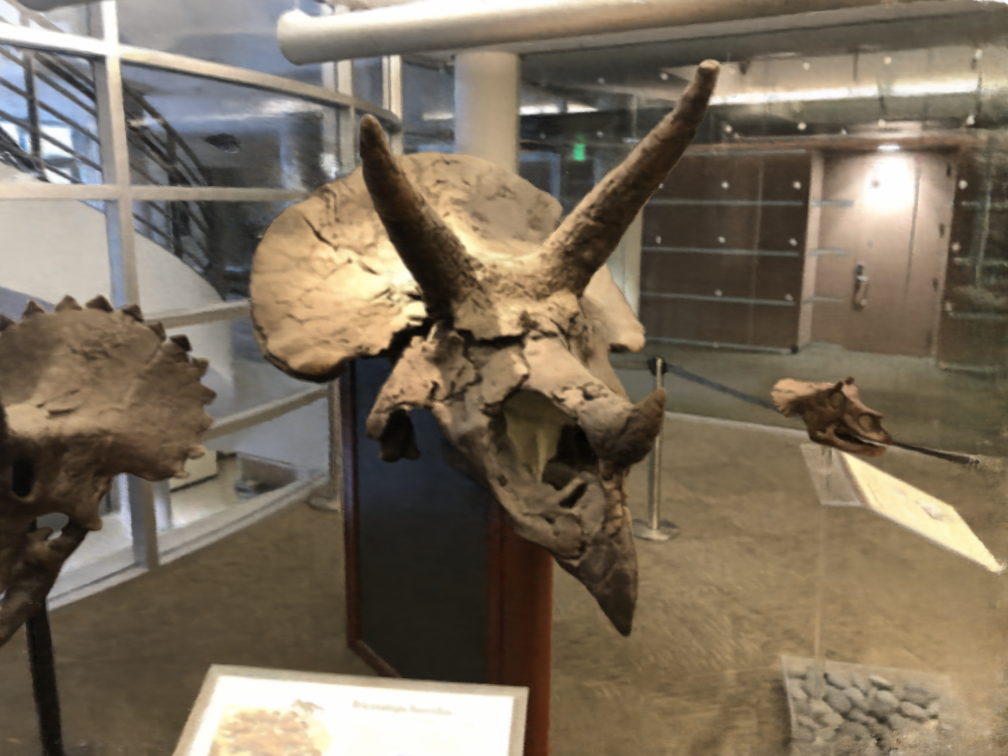}{NeRF} &
\crophorn{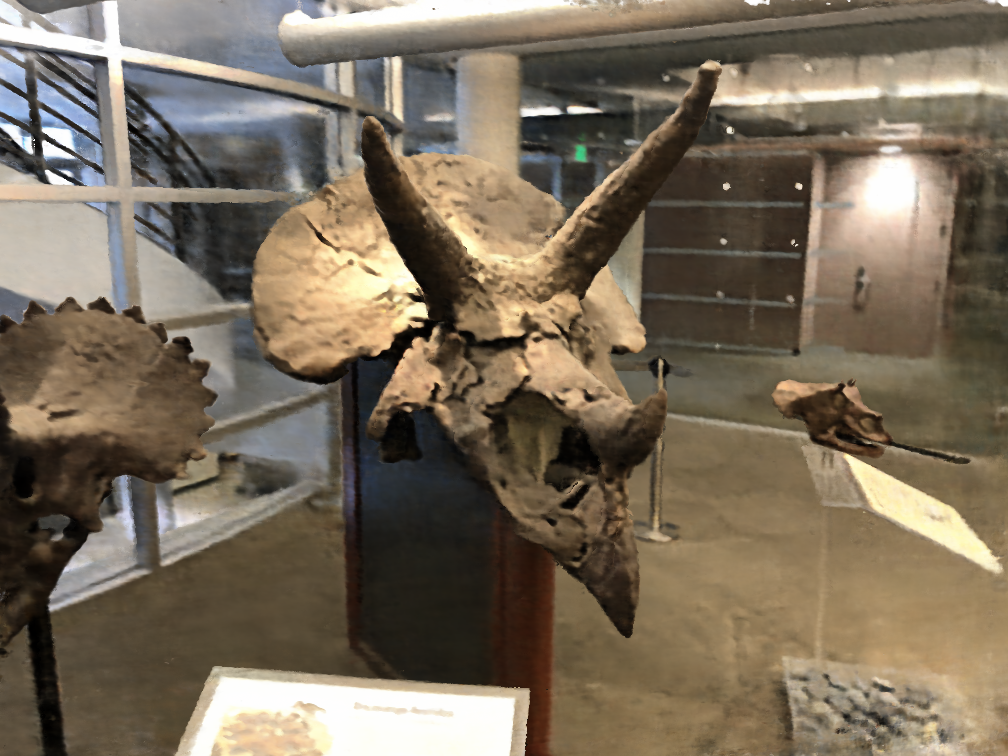}{NeRF-SR} &
\crophorn{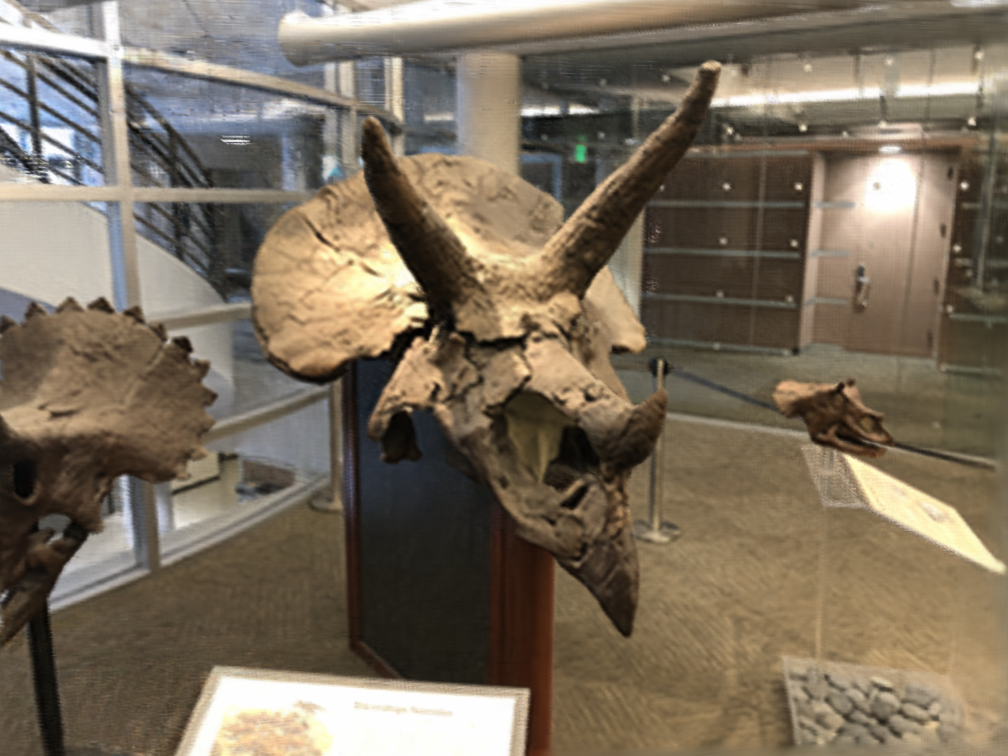}{TensoRF} &
\crophorn{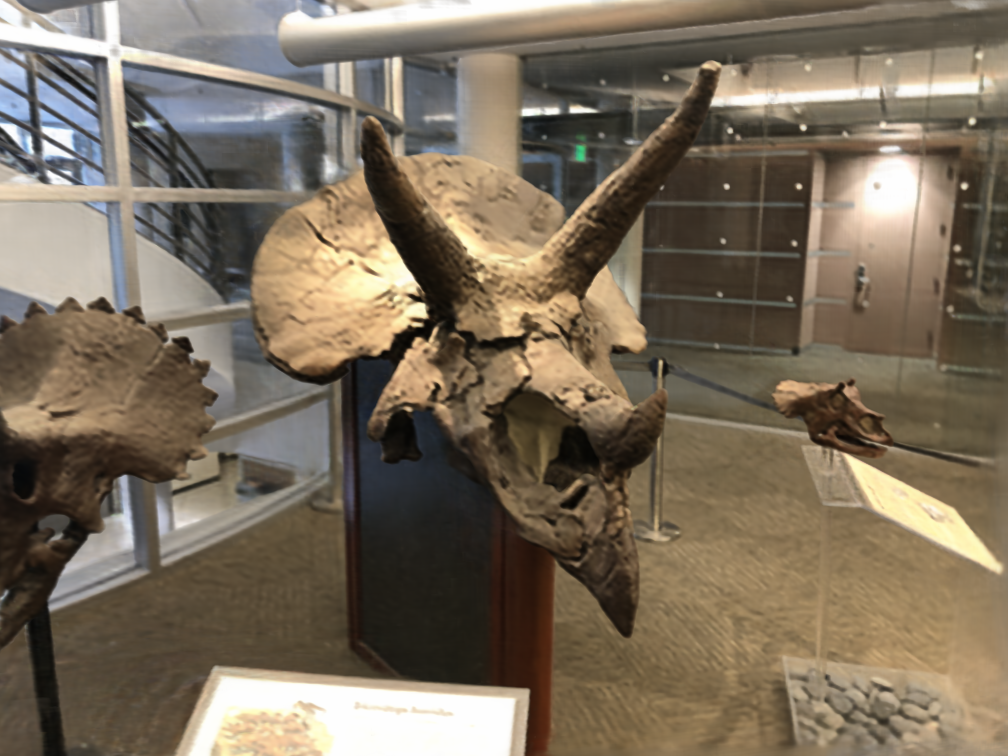}{Ours} \\

\makecell[c]{
\includegraphics[trim={0px 0px 0px 0px}, clip, width=\resultsfigwidth]{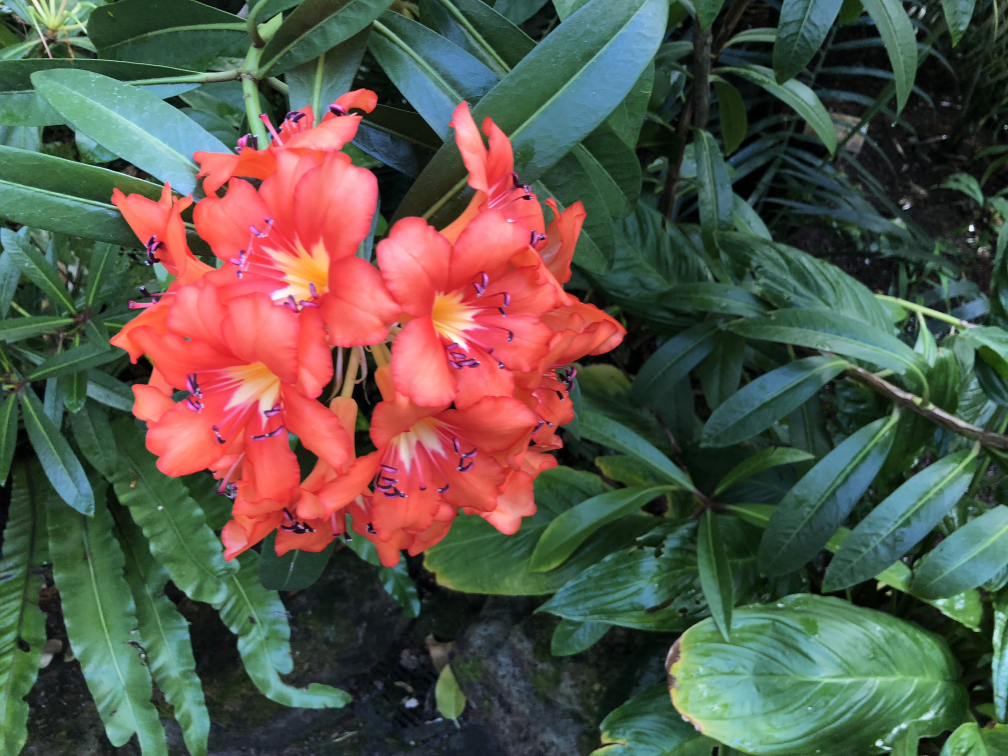}
\put(-60,35){{\color{red}{\huge$\square$}}}
\put(-103,63){{\color{green}{\huge$\square$}}} \\
}
&
\cropflower{photo/visual_llff/flower_gt.png}{Ground Truth} &
\cropflower{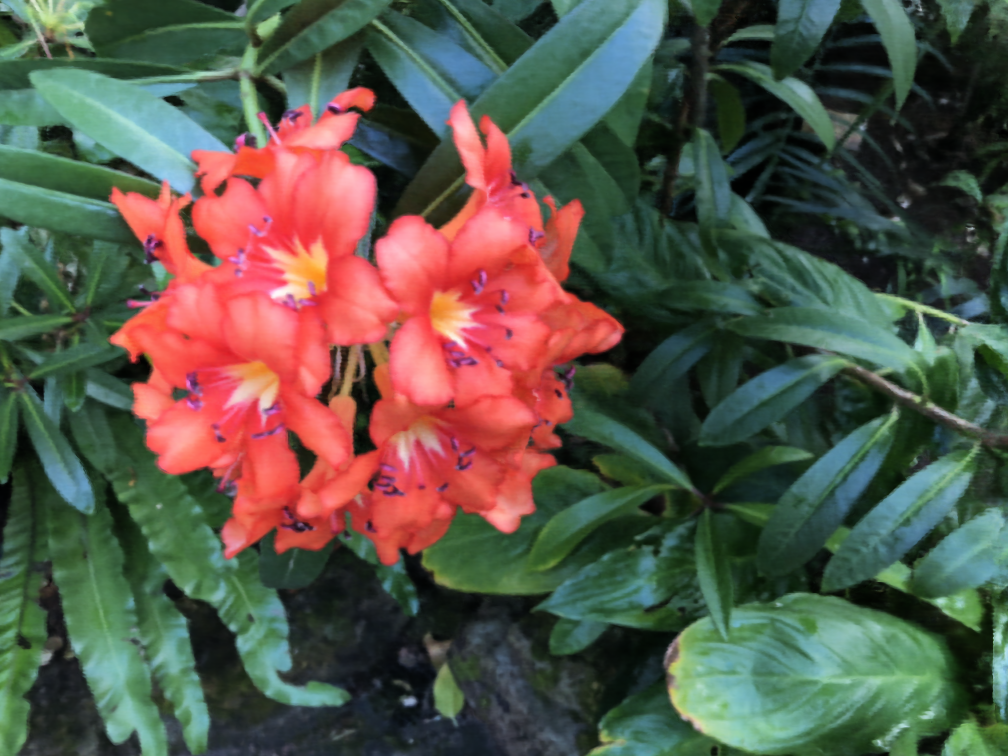}{NeRF} &
\cropflower{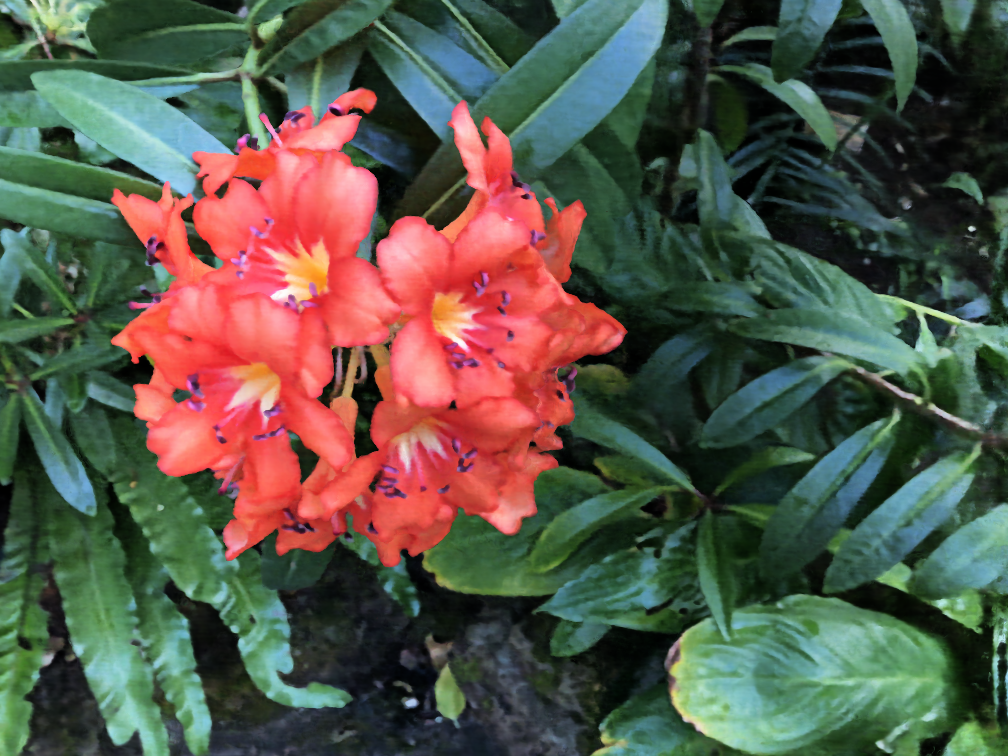}{NeRF-SR} &
\cropflower{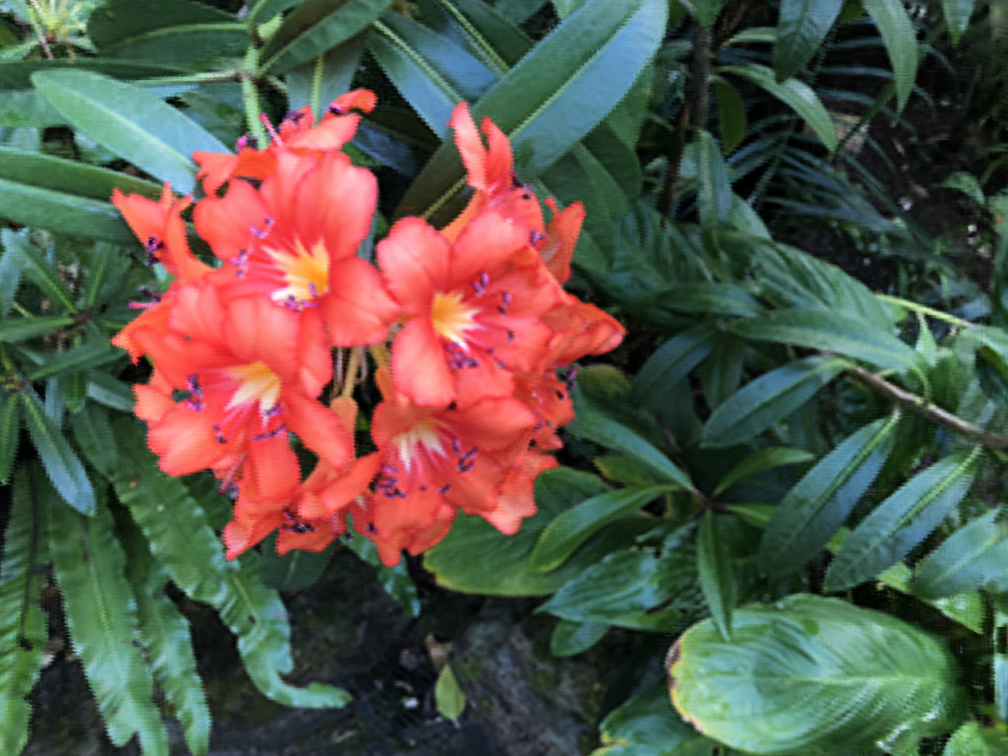}{TensoRF} &
\cropflower{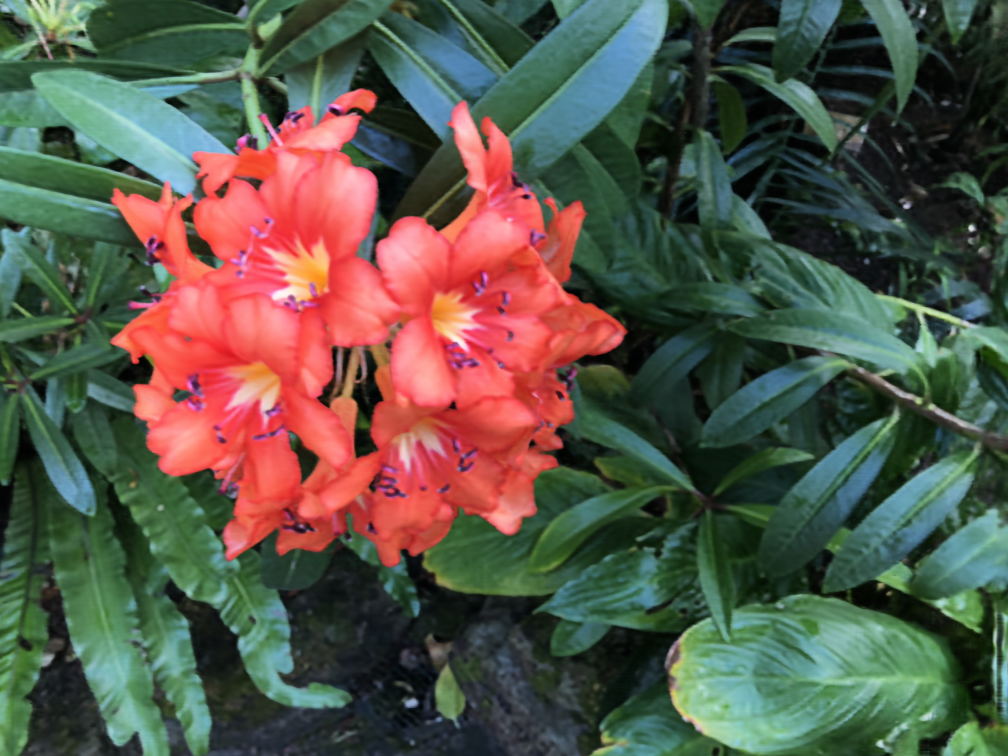}{Ours}


\end{tabular}
\vspace{-0.1cm}
\caption{Comparison of the synthesized super-resolution views (x4) of different methods on the LLFF dataset. Our results show clearer details than NeRF, NeRF-SR, and TensoRF, such as the textures of leaves and flowers.} 
\label{visual_llff}
\vspace{-0.5cm}
\end{figure*}

\subsection{Evaluation}

In this section, we rely on comparative study to show the performance of our approach by using the following baseline and state-of-the-art methods:
\begin{itemize}
    \item{\textbf{NeRF \cite{NeRF}.} Leveraging NeRF’s implicit function representation, which enables new view synthesis at any resolution, we utilize publicly available NeRF code. This allows us to implement LR inputs and directly render HR images.}
    \item{\textbf{TensoRF \cite{TensoRF}.} We compare with TensoRF implementation of NeRF by using the official source code, which acts as the baseline of our method.}
    \item{\textbf{NeRF-SR \cite{NeRF-SR}.} NeRF-SR is designed for synthesizing HR novel views from LR image inputs. Our experimental data is derived from both the NeRF-SR source code and related literature.}
    \item{\textbf{NVSR \cite{NVSR}.} As a contemporary work to NeRF-SR, NVSR synthesizes high-resolution views through super-resolution tri-planes. To save training time, We use four scenes from the Blender dataset to train NVSR and the remaining four for testing. The official source code is used.}
    \item{\textbf{SISR based methods \cite{ZSSR,Swinir}.} We also compare our method with SISR-based methods by performing pre-processing or post-processing. Two representative SISR models are used: ZSSR \cite{ZSSR} and SWINIR \cite{Swinir}. For pre-processing: SISR is used to preprocess the LR ground truth images to obtain HR images for training Super-Resolution TensoRF: ZSSR-TensorRF, SWINIR-TensorRF. For post-processing: SISR is used to post-process the rendered LR views of TensoRF: TensoRF-ZSSR and ZSSR-TensorRF.}
\end{itemize}

We qualitatively evaluate the synthesized view quality against the ground truth under identical poses. We quantitatively evaluate our approach with the following three metrics: Peak Signal Noise Ratio (PSNR), Structural Similarity Index Measure (SSIM) \cite{SSIM}, and Learned Perceptual Image Patch Similarity (LPIPS) \cite{Perc}.

\begin{table*}[h]
\centering
\caption{Results of novel view synthesis on blender and LLFF datasets for scale factors ×2 and ×4 on four input resolutions: blender x2 (400x400$\rightarrow$800x800), blender x4 (200x200$\rightarrow$800x800), LLFF x2 (504x378$\rightarrow$1008x756), LLFF x4 (252x189$\rightarrow$1008x756).}
\label{table 1}
\begin{tabular}{c|c|cccl|cccl}
\hline
\multirow{2}{*}{\textbf{Method}} & \multirow{2}{*}{\textbf{Factor}} & \multicolumn{4}{c|}{\textbf{Blender}}                                               & \multicolumn{4}{c}{\textbf{LLFF}}                                                   \\ \cline{3-10} 
                                 &                                  & PSNR $\uparrow$ & SSIM $\uparrow$ & \multicolumn{2}{c|}{$\mathrm{LPIPS}\downarrow$} & PSNR $\uparrow$ & SSIM $\uparrow$ & \multicolumn{2}{c}{$\mathrm{LPIPS}\downarrow$} \\ \hline
NeRF                             &                                  & 28.06                & 0.923               & \multicolumn{2}{c|}{0.052}                           & 24.57                &  0.750               & \multicolumn{2}{c}{0.192}                            \\
TensoRF                          & \multirow{2}{*}{x2}              & 31.45                 & 0.952                & \multicolumn{2}{c|}{0.049}                           & 25.88                 & 0.810                & \multicolumn{2}{c}{0.175}                            \\
NeRF-SR                          &                                  & 30.08                & 0.939                & \multicolumn{2}{c|}{0.050}                           & 25.26                & 0.755               & \multicolumn{2}{c}{0.220}                            \\
Ours                             &                                  &\textbf{31.93}                 & \textbf{0.954}                & \multicolumn{2}{c|}{\textbf{0.042}}                           & \textbf{26.49}                & \textbf{0.837}                & \multicolumn{2}{c}{\textbf{0.155}}                            \\ \hline
NeRF                             &                                  & 27.47           & 0.910           & \multicolumn{2}{c|}{0.128}                      & 21.69           & 0.626           & \multicolumn{2}{c}{0.313}                       \\
TensoRF                          & \multirow{2}{*}{x4}              & 28.01           & 0.910           & \multicolumn{2}{c|}{0.113}                      & 23.82           & 0.694           & \multicolumn{2}{c}{0.358}                       \\
NeRF-SR                          &                                  & 28.46           & 0.921           & \multicolumn{2}{c|}{0.076}                      & 23.51           & 0.693           & \multicolumn{2}{c}{0.297}                       \\
Ours                             &                                  & \textbf{29.69}  & \textbf{0.929}  & \multicolumn{2}{c|}{\textbf{0.069}}             & \textbf{24.80}  & \textbf{0.749}  & \multicolumn{2}{c}{\textbf{0.283}}              \\ \hline
\end{tabular}
\end{table*}

\begin{figure*}[htb]
\centering
\scriptsize
\begin{tabular}{cccc}

\makecell[c]{
\includegraphics[trim={0px 0px 0px 0px}, clip, width=\resultsfigwidthablego]{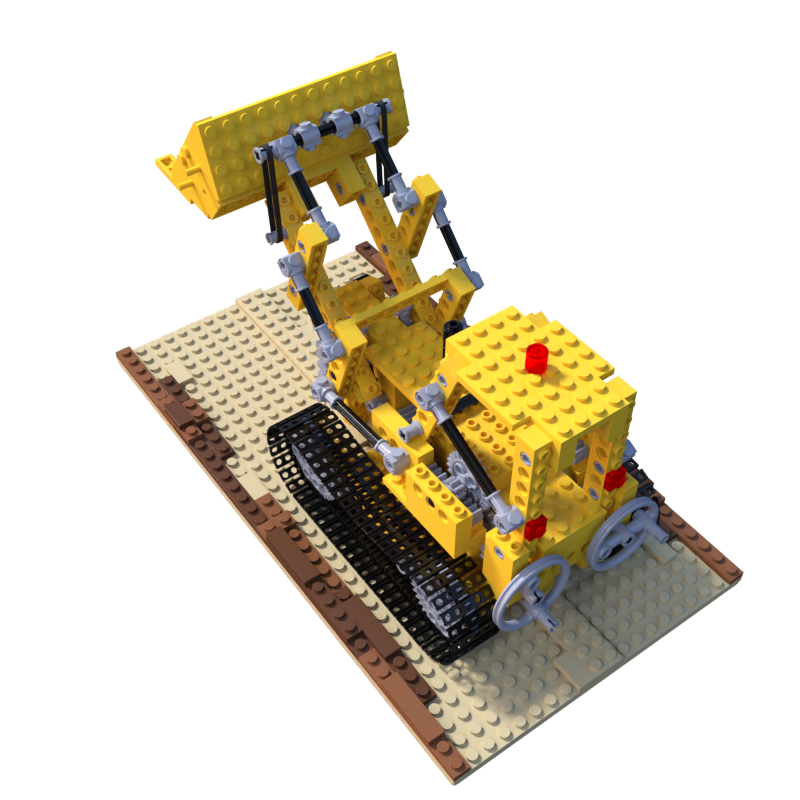}
    \put(-66,68){{\color{red}{\huge$\square$}}}
    \put(-57,8){{\color{green}{\huge$\square$}}}
     \\
}
&

\croplegoab{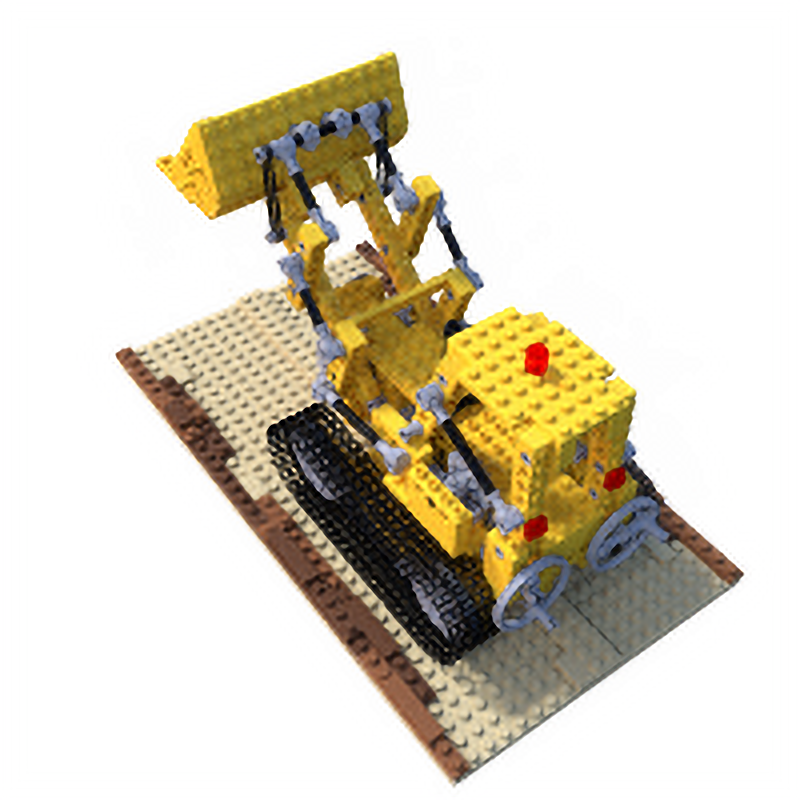}{TensoRF-ZSSR} &
\croplegoab{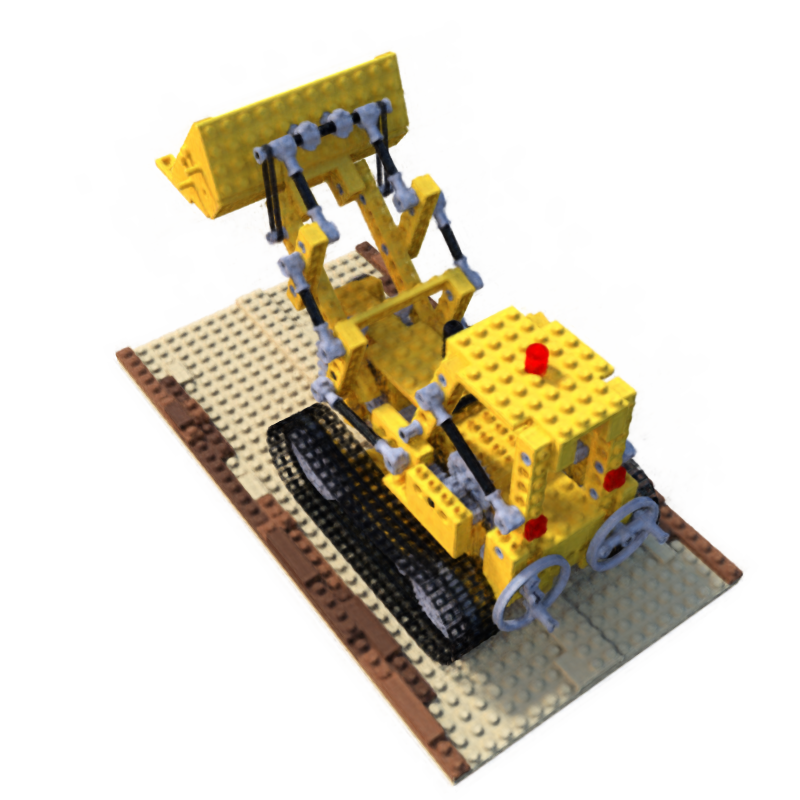}{ZSSR-TensoRF} &
\croplegoab{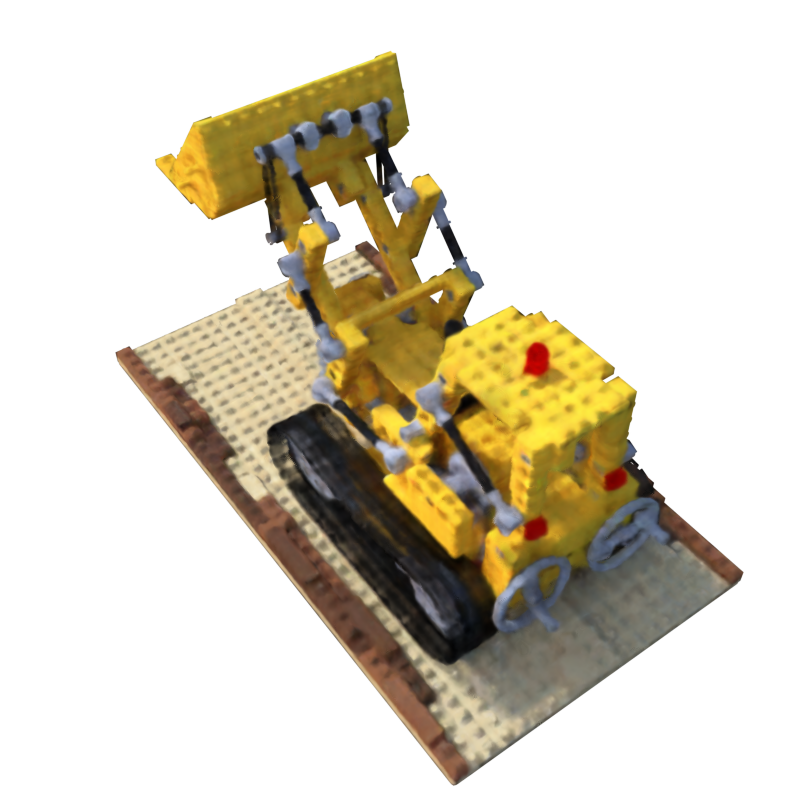}{NVSR} 
\\
\croplegoab{photo/visual_blender/lego_gt_190.png}{GT}&
\croplegoab{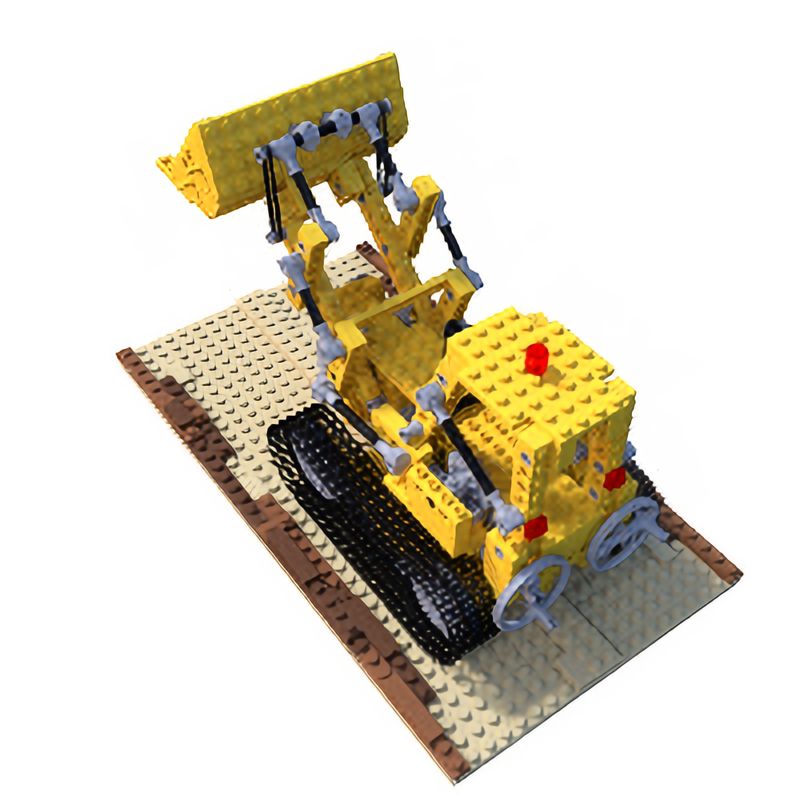}{TensoRF-Swinir} &
\croplegoab{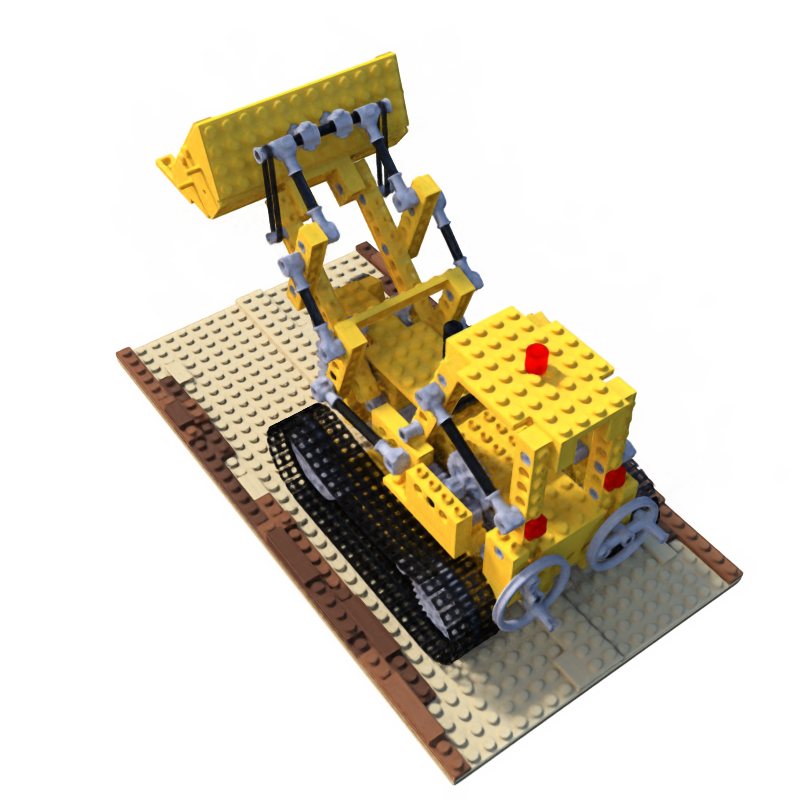}{Swinir-TensoRF} &
\croplegoab{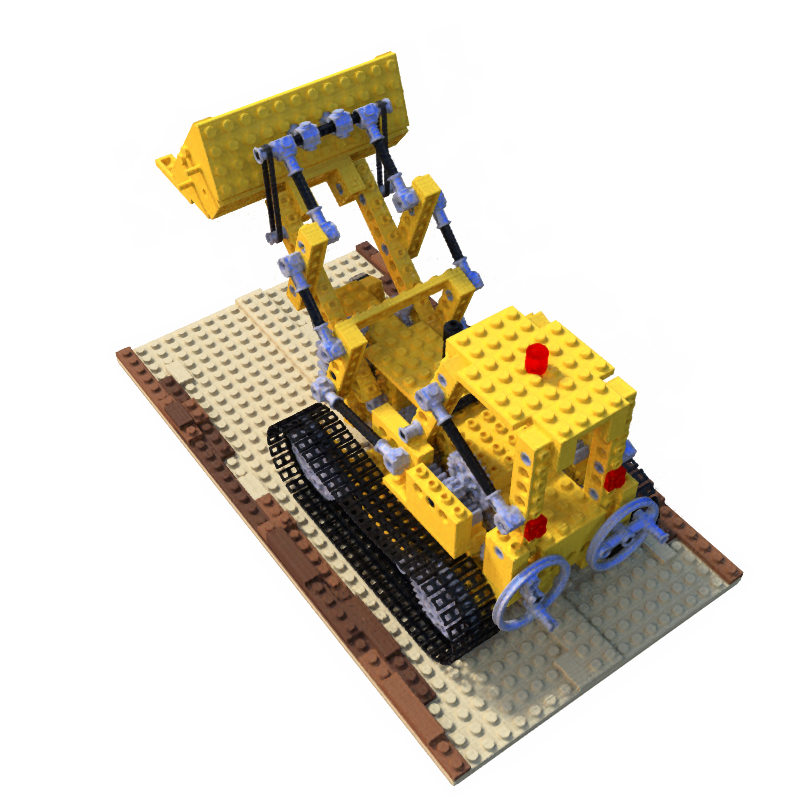}{Ours}
\end{tabular}
\caption{Visual comparison of our results with that of NVSR and SISR-based methods on Lego.} 
\label{2DSR}
\vspace{-0.5cm}
\end{figure*}

\subsection{Main Results}

\textbf{Qualitative evaluation.} 
Figure \ref{visual_blender} and \ref{visual_llff} visually compare super-resolution rendering results of different methods on both the Blender and LLFF datasets, where the input resolution is 200, and the upsampling factor is 4. NeRF and TensoRF suffer from obvious blurring effects due to the existence of a sampling gap. NeRF-SR can obtain better visual quality because the employment of the supersampling strategy can make up the sampling gap. In contrast, our results show more precise details that are closer to the ground truth. This can be contributed to the advantage of our parameterized SDM model which is superior than the linear interpolation method (i.e. average pooling) used in NeRF-SR. In Figure \ref{2DSR}, we visually compare our method with NVSR and the SISR-based methods. The results of NVSR, TensoRF-ZSSR and ZSSR-TensoRF suffer from the problems of blurry and structure distortion. Although the results of TensoRF-Swinir and Swinir-TensoRF perform better than those of ZSSR, they still suffer from the problem of structure distortion. Compared with pre-processing, the post-processing may easily confront the problem of structure distortions in image details. Instead, our results exhibit better visual quality because the optimization of our model fully happens in the 3D space supervised by our SDM model. Besides, our method also demonstrates its advantages on the training speed, the consumption of scene data and the quality of the rendered view in Figure \ref{introduction}. The speedup can be contributed to the employment of our coarse-to-fine optimization strategy, where the training cost for the lightweight SDM model is very low and can be neglected compared to training NeRF.

\begin{figure*}[htb]
\centering
\includegraphics[width=0.9\textwidth]{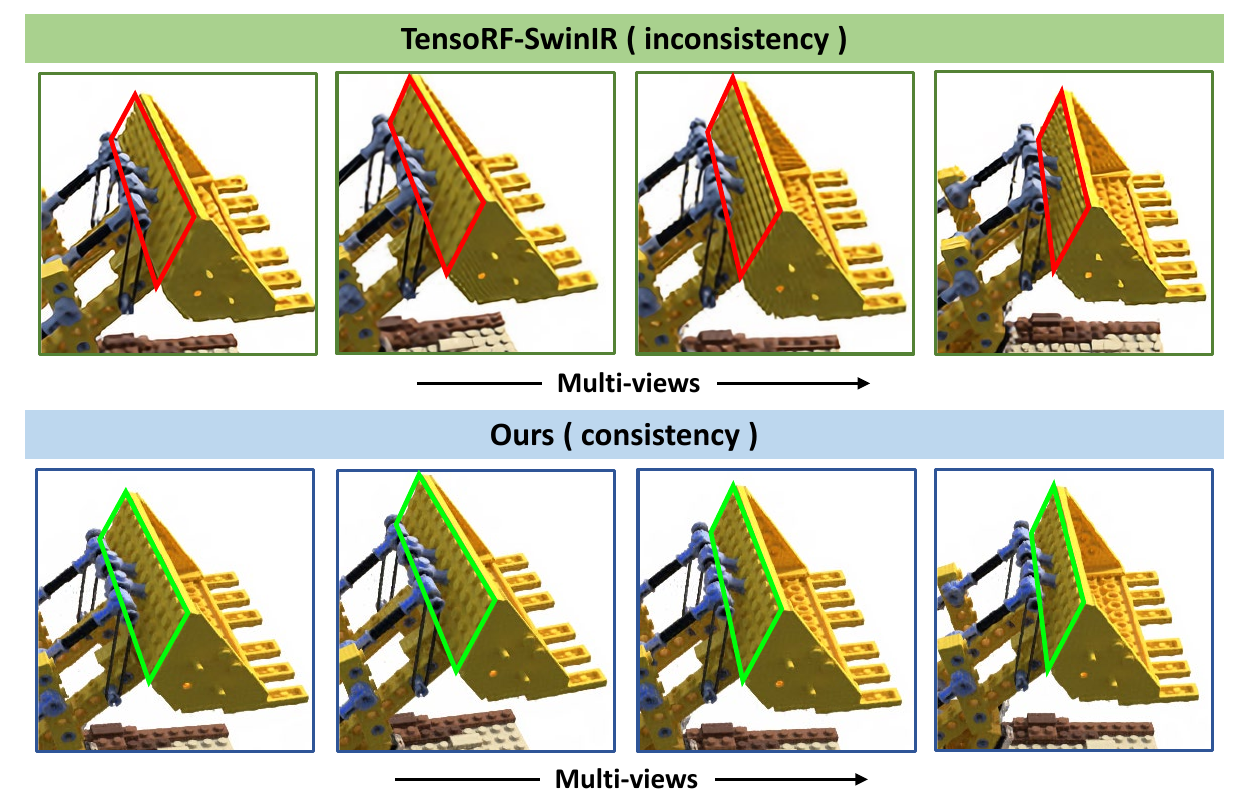}
\caption{Qualitative Comparison of Multi-view Consistency. The synthetic views of TensoRF-SwinIR suffer from inconsistent structure or texture (see the red boxes in the first row), whereas our model maintains remarkable multi-view consistency.}
\label{muti-view}
\vspace{-0.5cm}
\end{figure*}

\begin{table}[h]
\caption{Comparison of our results (×4) with that of NVSR and the SISR based methods: ZSSR-TensoRF, Swinir-TensoRF, TensoRF-ZSSR and TensoRF-Swinir.}
\label{table_2}
\begin{tabular}{l|ccc}
\hline
\multicolumn{1}{c|}{\textbf{Method}} & PSNR $\uparrow$ & SSIM $\uparrow$ & LPIPS $\downarrow$\\ \hline
TensoRF-ZSSR              & 27.93               & 0.894               & 0.124               \\
ZSSR-TensoRF              & 29.60               & 0.918               & 0.090               \\
TensoRF-Swinir              & 27.27               & 0.897               & 0.100               \\
Swinir-TensoRF              & 30.05               & 0.920               & 0.070               \\
NVSR              & 27.71               & 0.892               & 0.116               \\
Ours                  & \textbf{30.36}             &   \textbf{0.925}            &  \textbf{0.064}       \\ \hline
\end{tabular}
\end{table}

\textbf{Quantitative evaluation.} Table \ref{table 1} outlines the quantitative evaluation results. For the Blender dataset, we train the models on resolutions of 400 × 400 and 200 × 200 while testing them at 2x and 4x scales, respectively. Similarly, for the LLFF dataset, we train the models on resolutions of 504 × 378 and 252 × 189, and test them at 2x and 4x scales. It is easy to observe that our results exhibit the best performances on all the evaluation measures across the two datasets. In the x2 super-resolution task, tensor neural representation brings higher benefits than the dense radiation field. In the x4 super-resolution task, the sampling gap is larger, and the dense radiation field brings greater benefits. Due to differences in test scenarios, the results of NVSR are not included in Table \ref{table 1}. In Table \ref{table_2}, we compare our results with NVSR as well as that of the methods based on SISR under the same test configurations. It is obvious that our method outperforms the others with substantial improvements in PSNR, SSIM, and LPIPS metrics. The recent Swinir method outperforms ZSSR in super-resolution training. The post-processing is inferior to pre-processing, which is because that the super-resolution methods may easily lead to view-inconsistency without considering multi-view information.

\textbf{Multi-view consistency evaluation.} We demonstrate the view consistency through multi-view analysis. In Figure \ref{muti-view}, we compare our results with that of TensoRF-SwinIR. It is interesting to observe that SwinIR has disrupted the multi-view 3D consistency of TensorRF for super-resolution. For example, the structure or texture of the red boxes becomes distorted for consistent views. In contrast, our results can well preserve the multi-view consistency.

\subsection{Ablation Study}

In this section, we rely on ablation study to show the effectiveness of each component of method by adding one component each time starting from the baseline model TensoRF, including the super-resolution training, the SDM model, the gradient view, and the temporal ensemble. (a) \textit{+ super-resolution training} denotes performing super-resolution training of TensoRF as with that of NeRF-SR \cite{NeRF-SR}. (b) \textit{+ SDM} denotes performing super-resolution training of TensoRF supervised by using our SDM model but without gradient view, i.e., (a)+SDM. (c) \textit{+ gradient view} denotes the embedding of the gradient information in the SDM model for (b). (d) \textit{+temporal ensemble} denotes adding temporal ensemble for (c), which equals our entire model.

\begin{figure*}[htb]
\centering
\scriptsize
\begin{tabular}{cccccc}

\makecell[c]{
\includegraphics[trim={0px 0px 0px 0px}, clip, width=\resultsfigwidthab]{photo/visual_blender/ship_gt.png}
    \put(-47,33){{\color{red}{\huge$\square$}}}
     \put(-63,25){{\color{green}{\huge$\square$}}}
   \\
}
&
\cropshipab{photo/visual_blender/ship_TensoRF.png}{TensoRF} &
\cropshipab{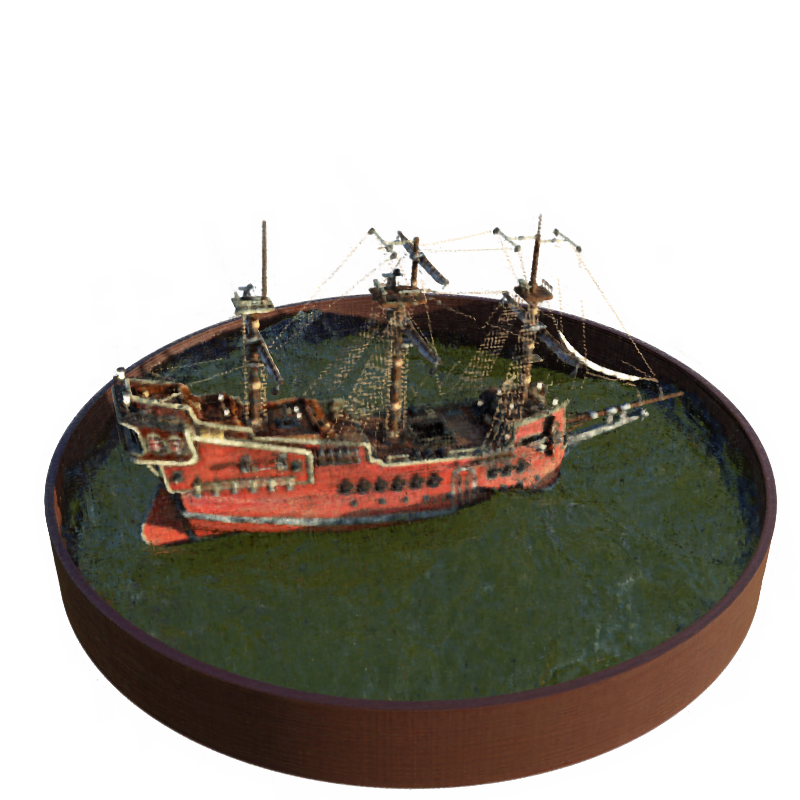}{+super-resolution training} &
\cropshipab{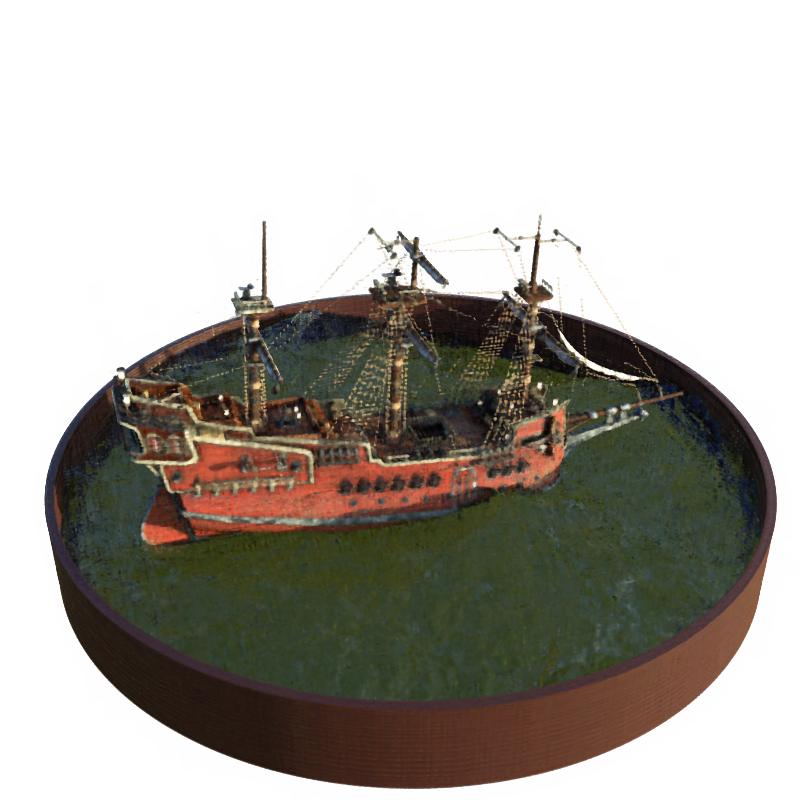}{+SDM} &
\cropshipab{photo/visual_blender/ship_ours.png}{Ours} &
\cropshipab{photo/visual_blender/ship_gt.png}{GT}
\end{tabular}
\caption{Demonstration of the visual ablation results with respect to super-resolution training, SDM and our final solution. Each of our methods can bring visual improvements to the rendered views.} 
\label{abstudy}
\end{figure*}

We first present the visual results in Figure \ref{abstudy}. It is easy to find that the super-resolution training can help to improve the image quality compared with the baseline TensoRF. SDM can help to improve the details (e.g. structure or texture) of the rendered view. The employment of a temporal ensemble would further improve the smoothness of the result. Then, in Table \ref{table_3}, we present the quantitative ablation results at upscale factor of x4.

\begin{table}[h]

\caption{Quantitative analysis of ablation study results at a scale factor of x4. }
\label{table_3}
\begin{tabular}{l|ccc}
\hline
\multicolumn{1}{c|}{\textbf{Method}} & PSNR $\uparrow$ & SSIM $\uparrow$ & LPIPS $\downarrow$\\ \hline
Baseline(TensoRF)               & 28.01               & 0.910               & 0.113               \\
(a) +super-resolution training   & 29.01              & 0.920              &  0.079        \\
(b) +SDM     & 29.51               & 0.926              &  0.073              \\
(c) +gradient view     & 29.59               & 0.927              &  0.070              \\
(d) +temporal ensemble                  & \textbf{29.69}             &   \textbf{0.929}            &  \textbf{0.069}       \\ \hline
\end{tabular}
\end{table}

\textbf{Effectiveness of the super-resolution training.} According to the results listed in Table \ref{table_3}, the super-resolution training can achieve a PSNR gain of 1.0dB, which can be contributed to the employment of the supersampling strategy to make up the sampling gap, where the average pooling mapping is used to perform supervised super-resolution training which is similar to that of NeRF-SR. Super-resolution training can improve the spatial sampling rate of the radiance field, allowing it to obtain high-quality novel views with high-resolution volume rendering.

\textbf{Effectiveness of the SDM mapping.} The results of (a) are still insufficient because the average operation for super-resolution training cannot support obtaining the fine details. After replacing the average pooling with our SDM model, our results (b) in Table \ref{table_3} can achieve a further improvement, such as the PSNR gain of 0.5dB. These improvements indicate that our SDM model can be used as a suitable supervisor for high-frequency super-resolution training.

\textbf{Effectiveness of gradient view and temporal ensemble.} As shown in Table \ref{table_3} (c) and (d), the employment of gradient view and temporal ensemble can also produce positive effects towards lifting the performance of our method. For example, they can totally achieve a 0.18dB improvement on PSNR score. The improvements can contribute to the regularization of gradient view information and the temporal ensemble strategy. 

Notice that our final solution has significantly improved the performance of the baseline (i.e. TensoRF) on all the evaluation items. Together with the visual comparison results, we have demonstrated the ability of each component of our approach.


\section{Conclusion}
In this paper, we propose a zero-shot super-resolution training method for neural radiance fields by focusing on addressing the problem of the lack of high-resolution ground truth. By employing the idea of internal learning, we can obtain a single-scene degeneration mapping model, which is further used in inverse rendering to supervise the super-resolution training of the fine NeRF for synthesizing high-quality, high-resolution novel views without consuming any external high-resolution scene data. With the temporal ensemble strategy, our method can compensate for the errors of super-resolution scene estimation. The comparative and ablation studies show that our method can achieve state-of-the-art performances by producing high-quality novel views with more fine details.

\section*{Acknowledgments}
This work was supported in part by the Zhejiang Provincial Natural Science Foundation of China (No. LY22F020028), the National Natural Science Foundation of China (No. 62372147), the Fundamental Research Funds for the Provincial Universities of Zhejiang under (No. GK229909299001-001), the National Natural Science Foundation of China (No. U21B2040, 62125201, 61972121), the Zhejiang Provincial Natural Science Foundation of China (No. LDT23F02025F02), the Open Project Program of the State Key Laboratory of CADCG(No. A2304), Zhejiang University, Aeronautical Science Foundation of China (No. 2022Z0710T5001).

\section*{Conflict of interest}
The authors declare that they have no conflict of interest.

\bibliographystyle{model1-num-names}
\newpage
\bibliography{cas-refs}

\end{document}